\documentclass{article}

\usepackage{microtype}
\usepackage{graphicx}
\usepackage{subfigure}
\usepackage{booktabs} 

\usepackage{hyperref}
\usepackage{threeparttable}

\usepackage{amsmath, amsthm, amssymb, mathtools}
\usepackage[ansinew]{inputenc}
\usepackage[algo2e]{algorithm2e}
\usepackage[usenames]{xcolor}
\usepackage{bm, xspace}
\usepackage{pifont}
\usepackage{multirow, hhline, array, graphicx, multicol}
\usepackage{tikz, pgfplots}
\usepackage{enumitem}
\usepackage{makecell}
\usepackage{fullpage}

\newtheorem{Definition}{Definition}
\newtheorem{Theorem}{Theorem}
\newtheorem{Lemma}{Lemma}
\newtheorem{Claim}{Claim}
\newtheorem{Assumption}{Assumption}

\usepackage{algorithm}
\usepackage{algorithmic}
\usepackage{natbib}

\renewcommand\arraystretch{1.9}
\long\def\comment#1{}
\begin{document}

\title{On Learning Sparsely Used Dictionaries from 
Incomplete Samples}



%
%
\author{Thanh V. Nguyen, Akshay Soni, and Chinmay Hegde 
\thanks{Email: \{thanhng, chinmay\}@iastate.edu;
  akshay@oath.com. T. N. and C. H. are with the Electrical and
  Computer Engineering Department at Iowa State University. A. S. is with Yahoo! Research. This work was supported in part by the National Science Foundation under grants CCF-1566281 and CCF-1750920.}
}

\maketitle

\vskip 0.3in





\def\1{\bm{1}}

\newcommand*{\bigcdot}{\bullet}

\newcommand{\cA}[1]{A_{\bigcdot #1}}
\newcommand{\cAi}{\cA{i}}
\newcommand{\cAj}{\cA{j}}
\newcommand{\cAl}{\cA{l}}
\newcommand{\cAS}{\cA{S}}

\newcommand{\cB}[1]{A_{\bigcdot #1}}
\newcommand{\cBi}{\cB{i}}
\newcommand{\cBS}{\cB{S}}

\newcommand{\rA}[1]{A_{#1 \bigcdot}}
\newcommand{\rAj}{\rA{j}}
\newcommand{\rAR}{\rA{R}}
\newcommand{\rAG}{\rA{\Gamma}}
\newcommand{\rAGa}{\rA{\Gamma_{\alpha}}}
\newcommand{\rAGap}{\rA{\Gamma_{\alpha'}}}
\newcommand{\rAl}{\rA{l}}
\newcommand{\aG}{a_{\Gamma}}

\newcommand{\AR}[1]{A_{R, #1}}
\newcommand{\AG}[1]{A_{\Gamma, #1}}

\newcommand{\cg}[1]{g_{\bigcdot #1}}
\newcommand{\cgi}{\cg{i}}
\newcommand{\cgS}{\cg{S}}
\newcommand{\gR}[1]{g_{R, #1}}

\newcommand{\PGami}[1]{\mathcal{P}_{\Gamma_{#1}}}

\def\ghat{\widehat{g}}
\def\Rhat{\widehat{R}}
\def\dhat{\widehat{d}}
\def\ehat{\widehat{e}}
\def\Muv{M_{u,v}}
\def\Mhuv{\widehat{M}_{u,v}}
\def\VarE{\mathcal{E}}
\def\Ne{N_{\varepsilon}}
\def\Rad{R}
\def\Aupb{\mathcal{A}}
\def\Bupb{\mathcal{B}}
\def\PGam{\mathcal{P}_{\Gamma}}
\def\nbar{\bar{n}}

\newcommand*{\Asmp}[1]{Assumption \textnormal{\textbf{A#1}}}
\newcommand*{\AsmpB}[1]{Assumption \textnormal{\textbf{B#1}}}

\def\sgn{\textnormal{sgn}}
\def\supp{\textnormal{supp}}
\def\diag{\textnormal{diag}}
\def\card{\textnormal{card}}
\def\thres{\textnormal{threshold}}
\def\polylog{\textnormal{polylog}}
\def\iid{\text{i.i.d.}}
\def\whp{\text{w.h.p.}}
\def\wrt{\text{w.r.t.}}
\def\ie{\text{i.e.}}
\def\eg{\text{e.g.}}

\newcommand{\ceil}[1]{\left \lceil #1 \right \rceil}

\def\cmark{\ding{51}}
\def\xmark{\ding{55}}

\newcommand{\E}{\mathbb{E}}
\newcommand{\Prob}{\mathbb{P}}
\newcommand{\sigmae}{\sigma_\varepsilon}

\newcommand{\R}{\mathbb{R}}

\DeclarePairedDelimiterX{\norm}[1]{\lVert}{\rVert}{#1} 
\DeclarePairedDelimiterX{\inprod}[2]{\langle}{\rangle}{#1, #2}
\DeclarePairedDelimiterX{\abs}[1]{\lvert}{\rvert}{#1}

\DeclarePairedDelimiterX{\bigO}[1]{(}{)}{#1}
\def\Otilde{\widetilde{O}}
\def\Omgtilde{\widetilde{\Omega}}

\newcommand{\sgnEvent}{\mathcal{F}_{x^*}}


\begin{abstract}

Most existing algorithms for dictionary learning assume that all entries of the (high-dimensional) input data are fully observed. However, in several practical applications (such as hyper-spectral imaging or blood glucose monitoring), only an incomplete fraction of the data entries may be available. For incomplete settings, no provably
correct and polynomial-time algorithm has been reported in the dictionary learning literature. In this paper, we provide provable approaches for learning -- from incomplete samples -- a family of dictionaries whose atoms have sufficiently ``spread-out'' mass. First, we propose a descent-style iterative algorithm that linearly converges to the true dictionary when provided a sufficiently coarse initial estimate. Second, we propose an initialization algorithm
that utilizes a small number of extra fully observed samples to produce such a coarse initial estimate. Finally, we theoretically analyze their performance and provide asymptotic statistical and computational guarantees.

\end{abstract}


\section{Introduction}
\label{intro}

\subsection{Motivation}

In this paper, we consider a variant of the problem of \emph{dictionary learning}, a widely used unsupervised technique for learning compact (sparse) representations of high dimensional data. At its core, the challenge in dictionary learning is to adaptively discover a basis (or dictionary) that can {sparsely} represent a given set of data samples with as little empirical representation error as possible. The study of sparse coding enjoys a rich history in image processing, machine learning, and compressive sensing~\citep{elad06-denoising, aharon06-ksvd, olshausen97-sc,candes05-decoding,elad2,
  lista,boureau2010learning}. While the majority of these
aforementioned works involved heuristics, several exciting recent
results~\citep{spielman12-exact, agarwal13-exact, agarwal14-learning,
  arora-colt14, arora-colt15, sun15-complete, bartlett-nips17, nguyen-aaai18} have established
rigorous conditions under which their algorithms recover the
\emph{true} dictionary provided the data obeys a suitable generative model.
  
 An important underlying assumption that guides the success of all existing dictionary learning algorithms is the availability of (sufficiently many) data samples that are fully observed. Our focus, on the other hand, is on the special case where \emph{the given data points are only partially observed}, that is, we are given access to only a small fraction of the coordinates of the data samples. 
 
 Such a setting of partial/incomplete observations is natural in many applications like image-inpainting and demosaicing~\citep{elad2}. For example, this routinely appears in hyper-spectral imaging~\citep{xing-siam12} where entire spectral bands of signals could be missing or unobserved. Moreover, in other applications, collecting fully observed samples can be expensive (or in some cases, even infeasible). Examples include the highly unreliable continuous blood glucose (CBG) monitoring systems that suffer from signal dropouts, where often the task is to learn a dictionary from such incompletely observed signals~\citep{naumova-17}. 

Earlier works that tackle the incomplete variant of the dictionary learning problem only offer heuristic solutions~\citep{xing-siam12,naumova-17} or involve constructing intractable statistical estimators~\citep{soni-sfm16}. Indeed, the recovery of the true dictionary involves analyzing an extremely non-convex optimization problem that is, in general, not solvable in polynomial time~\citep{loh-nips11}. To our knowledge, our work is the first to give a theoretically sound as well as tractable algorithm to recover the exact dictionary from missing data (provided certain natural assumptions are met). 

\subsection{Our Contributions}

In this paper, we make concrete theoretical algorithmic progress to the dictionary learning problem with incomplete samples.  Inspired by recent algorithmic advances in dictionary learning~\citep{arora-colt14,arora-colt15}, we adopt a learning-theoretic setup. Specifically, we 
assume that each data sample is synthesized from a generative model with an unknown dictionary and a random \emph{$k$-sparse} coefficient vector (or sparse code). Mathematically, the data samples $Y = [y^{(1)}, y^{(2)}, \ldots, y^{(p)}] \in \mathbb{R}^{n \times p}$ are of the form
$$
Y = A^* X^* \, ,
$$
where $A^* \in \mathbb{R}^{n \times m}$ denotes the dictionary and $X^* \in \mathbb{R}^{m \times p}$ denotes the (column-wise) $k$-sparse codes. 

However, we do not have direct access to the data; instead, each high-dimensional data sample is further subsampled 
such that only a small fraction of the entries are observed. The assumption we make is that each entry of $Y$ is observed independently with probability $\rho \in (0, 1]$. 
For reasons that will become clear, we also assume that the ground truth dictionary $A^*$ is both \emph{incoherent} (i.e., the columns of $A^*$ are sufficiently close to orthogonal) and \emph{democratic} (i.e., the energy of each atom is well spread). Both these assumptions are standard in the compressive-sensing literature. We clarify the generative model more precisely in the sequel.

Given a set of such (partially observed) data samples, our goal is to recover the true dictionary $A^*$.  Towards this goal, we make the following contributions: 

\begin{enumerate}[wide, labelwidth=!, labelindent=0pt]

\item Let us assume, for a moment, that we are given a coarse estimate
  $A^0$ that is sufficiently close to the true dictionary. We devise a
  descent-style algorithm that leverages the given incomplete data to
  iteratively refine the dictionary estimate; moreover, we show that
  it converges rapidly to an estimate within a small ball of the
  ground truth $A^*$ 
  (whose radius decreases given more samples). Our result can be informally summarized as follows:

\begin{Theorem}[Informal, descent]
  When given a ``sufficiently-close'' initial estimate $A^0$, there exists an iterative gradient descent-type algorithm that linearly converges to the true dictionary with $O(mk~\polylog(n))$ incomplete samples.
\end{Theorem}

Our above result mirrors several recent results in non-convex learning
that all develop a descent algorithm which succeeds given a good
enough initialization~\citep{sparsepca, wirtinger,procrustes}. Indeed, similar guarantees for descent-style algorithms (such as alternating minimization) exist for the related problem of \emph{matrix completion}~\citep{jain13-lowrank}, which coincides with our setting if $m \ll n$. However, our setting is distinct, since we are interested in learning \emph{overcomplete} dictionaries, where $m > n$. 


\item Having established the efficiency of the above refinement procedure, we then address the challenge of actually coming up with a coarse estimate of $A^*$. We do not know of a provable procedure that produces a good enough initial estimate using partial samples. To circumvent this issue, we assume availability of $O(m)$ \emph{fully} observed samples along with the partial samples\footnote{While this might be a limitation of our analysis, we emphasize that the number of full samples needed by our method is relatively small. Indeed, the state-of-the-art approach for dictionary learning~\citep{arora-colt15}
 requires $O(mk~\polylog(n))$ fully observed samples, while our method needs only $O(m~\polylog(n))$ samples, which represents a polynomial
  improvement since $k$ can be as large as $\sqrt{n}$. 
  }. 
Given this setting, we show that we can provide a ``sufficiently close'' initial estimate in polynomial time. Our result can be summarized as follows:

\begin{Theorem}[Informal, initialization]
  There exists an initialization algorithm that, given $O(m~\polylog(n))$ fully observed samples and an additional $O(mk~\polylog(n))$ partially observed samples, returns an initial estimate $A^0$ that is sufficiently close to $A^*$ in a column-wise sense.
\end{Theorem}
 
\end{enumerate}


\subsection{Techniques}

The majority of our theoretical contributions are fairly technical, so for clarity, we provide some non-rigorous intuition. 

At a high level, our approach merges ideas from two main themes in the
algorithmic learning theory literature. We build upon recent seminal,
theoretically-sound algorithms for sparse coding (specifically, the
framework of~\citet{arora-colt15}). Their approach consists of a
descent-based algorithm performed over the surface of a suitably
defined loss function of the dictionary parameters. The descent is 
achieved by alternating between updating the dictionary estimate and updating the sparse codes of the data samples. The authors prove that this algorithm succeeds provided that the codes are sparse enough, the columns of $A^*$ are incoherent, and that we are given sufficiently many samples. 

However, a direct application of the above framework to the partially observed setting does not seem to succeed. To resolve this, we leverage a specific property that is commonly assumed in the matrix completion literature: we suppose that the dictionaries are not ``spiky'' and that the energy of each atom is spread out among its coordinates; specifically, the \emph{sub}-dictionaries formed by randomly sub-selecting rows are still incoherent. We call such dictionaries \emph{democratic}, following the terminology of~\citet{davenport-democ09}. (In matrix completion papers, this property is also sometimes referred to incoherence, but we avoid doing so since that overloads the term.)
Our main contribution is to show that democratic, incoherent dictionaries can be learned via a similar alternating descent scheme if only a small fraction of the data entries are available. Our analysis is novel and distinct than that provided in~\citep{arora-colt15}.

Of course, the above analysis is somewhat local in nature since we are using a descent-style method. In order to get global guarantees for recovery of $A^*$, we need to initialize carefully. Here too, the spectral initialization strategies suggested in earlier dictionary learning papers~\citep{arora-colt14,arora-colt15} do not succeed. To resolve this, we again appeal to the democracy property of $A^*$. We also need to assume that provided a small hold-out set of additional, \emph{fully} observed samples is available\footnote{We do not know how to remove this assumption, and it appears that techniques stronger than spectral initialization (e.g., involving higher-order moments) are required.}. Using this hold-out set (which can be construed as additional prior information or ``side'' information) together with the available samples gives us a spectral initialization strategy that provably gives a good enough initial estimate.

Putting the above two pieces together: if we are provided
$\Omega(mk/\rho^4~\polylog~n)$ partially observed samples using the
aforementioned generative model, together with an additional
$\Omega(m~\polylog~n)$ full samples, then we can guarantee a fast, provably accurate algorithm to estimate $A^*$. See Table \ref{tbl_overview} for a summary of our results, and comparison with existing work. We remark that while our algorithms only succeed up to sparsity level $k \leq O(\rho\sqrt{n})$, we obtain a running time improvement over the best available dictionary learning approaches.

Most theorems that we introduce in the main paper are stated in terms of expected value bounds (i.e., we assume that infinitely many samples were given). Our finite-sample bounds (and consequently, the above sample complexity results) are derived using somewhat-tedious concentration arguments that are relegated to the appendix due to space constraints.

\begingroup
\renewcommand*\arraystretch{1.5} 

\begin{table*}[ht]
  \begin{center}
 \resizebox{\textwidth}{!}{%
 \begin{threeparttable}
\caption{Comparisons between different approaches.\label{tbl_overview}}
 
\begin{tabular}{|c|c|c|c|c|c|}
	\hline
	{Setting} & Reference & {\makecell{Sample complexity \\ w/o noise}} &
                                                                {Running time} & {Sparsity} & {\makecell{Incomplete samples}} \\
	\hline \hline
	\multirow{3}{*}{Regular} & \citep{spielman12-exact} & $O(n^2 \log n)$ & $\Omgtilde(n^4)$ & $O(\sqrt{n})$ & \xmark \\
	\hhline{~-----}
	& \citep{arora-colt14} & $\Otilde(m^2/k^2)$ &  $\Otilde(np^2)$ & $O(\sqrt{n})$ & \xmark \\
	\hhline{~-----}
	& \citep{arora-colt15} &  $\Otilde(mk)$ & $\Otilde(mn^2p)$ & $O(\sqrt{n})$ & \xmark \\
	\hline \hline
	\multirow{4}{*}{Incomplete} & ~\citep{xing-siam12} & \xmark & \xmark & \xmark & \cmark \\
	\hhline{~-----}
	& \citep{naumova-17} & \xmark  & \xmark & \xmark & \cmark \\
	\hhline{~-----}
	& This paper & \makecell{$\Otilde(mk/\rho^4)$ partial samples \\ $\Otilde(m)$ full samples}  &
                                                             $\Otilde(\rho mn^2p)$ & $O(\rho\sqrt{n})$ & \cmark \\
	\hline
\end{tabular}
{\small\sl
    \begin{tablenotes}
    \item \xmark\,  indicates no complexity guarantees. Here, $n$ is the data dimension; $m$ is the size of dictionary; $k$ is the sparsity of $x$; $p$ is the number of observed samples; $\rho$ is the subsampling probability.
    \end{tablenotes}}
    \end{threeparttable}
    }
    \end{center}  
\end{table*}

\endgroup

\subsection{Relation to Prior Work}

The literature on dictionary learning (or sparse coding) is very vast and hence our references to prior work will necessarily be incomplete; we refer to the seminal work of~\citet{elad2} for a list of applications. 
Dictionary learning with incompletely observed data, however, is far less well-understood. Initial attempts in this direction~\citep{xing-siam12} involve Bayesian-style techniques; more recent attempts have focused on alternating minimization techniques, along with incoherence- and democracy-type assumptions akin to our framework~\citep{naumova-schnass17, naumova-17}. However, none of these methods provide rigorous polynomial-time algorithms that provably succeed in recovering the dictionary parameters.

Our setup can also be viewed as an instance of matrix completion, which has been a source of intense interest in the machine learning community over the last decade~\citep{mc1,mc2}. The typical assumption in such approaches is that the data matrix $Y = A^* X^*$ is low-rank (i.e., $A^*$ typically spans a low-dimensional subspace). This assumption leads to either feasible convex relaxations, or a bilinear form that can be solved approximately via alternating minimization. However, our work differs significantly from this setup, since we are interested in the case where $A^*$ is over-complete; moreover, our guarantees are not in terms of estimating the missing entries of $Y$, but rather obtaining the atoms in $A^*$. Note that our generative model also differs from the setup of \emph{high-rank} matrix completion~\citep{highrank}, where the data is sampled randomly from a finite union-of-subspaces. In contrast, our data samples are synthesized via sparse linear combinations of a given dictionary.

In the context of matrix-completion, perhaps the most related work to ours is the statistical analysis of matrix-completion under the \emph{sparse-factor model} of~\citet{soni-sfm16}, which employs a very similar generative data model as ours. (Similar sparse-factor models have been studied in the work of~\citet{sparfa}, but again no complexity guarantees are provided.)
For this model, \citet{soni-sfm16} propose a highly non-convex statistical estimator for estimate $Y$ and provide error bounds for this estimator under various noise models. However, they do not discuss an efficient algorithm to realize that estimator. In contrast, we provide rigorous polynomial time algorithms, together with error bounds on the estimation quality of $A^*$. Overall, we anticipate that our work can shed some light on the design of provable algorithms for matrix-completion in such more general settings.

\subsection{Organization}

The remainder of the paper is organized as follows. Section~\ref{model} introduces some key definitions, a generative model for our data samples, and various assumptions about the model. Section~\ref{descent} introduces our main gradient-descent based algorithm, together with analysis. Section~\ref{init} introduces the initialization procedure for the descent algorithm. Section~\ref{exp} provides some representative numerical benefits of our approach. All proofs are deferred to the appendix unless stated explicitly.


\section{Preliminaries}
\label{model}

\emph{\bf Notation.}
Given a vector $x \in\R^{m}$ and a subset $S \subseteq [m]$, we
denote $x_S \in \R^m$ as a vector which equals $x$ in indices belonging to $S$ and equals zero elsewhere. We use $\cAi$ and $\rAj^T$ respectively to denote the $i^{\textrm{th}}$ column and the
$j^{\textrm{th}}$ row of matrix $A \in \R^{n\times m}$. We use $\cAS$ as the
submatrix of $A$ with columns in $S$. In contrast, we use $\rAG$ to
indicate the submatrix of $A$ with rows not in $\Gamma$ set to zero. 
Let $\supp(x)$ and $\sgn(x)$ 
 be the support and element-wise sign of $x$. Let $\mathrm{threshold}_{K}(x)$
be the \emph{hard-thresholding} operator that sets all entries of $x$ with magnitude less than $K$ to zero. The symbol $\norm{\cdot}$ refers to the $\ell_2$-norm, unless otherwise specified.

For asymptotic analysis, we use $\Omgtilde(\cdot)$ and $\Otilde(\cdot)$ to represent $\Omega(\cdot)$ and $O(\cdot)$ up to (unspecified) poly-logarithmic factors depending on $n$. Besides, $g(n)=O^*(f(n))$ denotes $g(n)\le Kf(n)$ for some sufficiently small constant $K$.
Finally, the terms ``with high probability'' (abbreviated to \whp) is
used to indicate an event with failure probability
$O(n^{-\omega(1)})$. We make use of the following definitions.

\begin{Definition}[Incoherence]
  \label{def_incoherence}
  The matrix $A$ is \emph{incoherent} with parameter $\mu$ if the following holds
  for all columns $i \neq j$:
    $$\frac{\abs{\inprod{\cAi}{\cAj}}}{\norm{\cAi}\norm{\cAj}} \leq
    \frac{\mu}{\sqrt{n}}.$$
\end{Definition}

The incoherence property requires the columns of $A$ to be approximately
orthogonal, and is a canonical property to resolve identifiability issues in dictionary
learning and sparse recovery. We distinguish this from the conventional notion of ``incoherence'' widely used in the matrix completion literature. This notion is related to a notion that we call \emph{democracy}, which we define next.  

\begin{Definition}[Democracy]
  Suppose that the matrix $A$ is $\mu$-incoherent. $A$ is further said to be
  \emph{democratic} if the submatrix $\rA{\Gamma}$ is $\mu$-incoherent
  for any subset $\Gamma \subset [n]$ of size $ \sqrt{n} \leq
  \abs{\Gamma} \leq n$.
\end{Definition}

This property tells us that the rows of $A$ have roughly the same
amount of ``information'', and that the submatrix of $A$ restricted to
any subset of rows $\Gamma$ is also incoherent. A similar concept (stated in terms of the restricted
isometry property) is well-known in the compressive sensing literature~\citep{davenport-democ09}. Several probabilistic constructions of dictionaries satisfy this property; 
typical examples include random matrices drawn from i.i.d.\ Gaussian or
Rademacher distributions. The $\sqrt{n}$ lower bound on $\abs{\Gamma}$
is to ensure that the submatrix of $A$ including only the rows in $\Gamma$ 
is balanced in terms of dimensions. 

We seek an algorithm that provides a provably ``good'' estimate of $A^*$. For this, we need a suitable measure of ``goodness''. The following notion of distance records the maximal column-wise difference between any estimate $A$ and $A^*$ in $\ell_2$-norm under a suitable permutation and sign flip.

\begin{Definition}[$(\delta, \kappa)$-nearness]
 The matrix $A$ is said to be $\delta$-close to $A^*$ if $\norm{\sigma(i)
   \cA{\pi(i)} - \cAi^*} \leq \delta$ holds for every $i = 1, 2,
 \dots, m$ and some
 permutation $\pi : [m] \rightarrow [m]$ and sign flip $\sigma : [m] :
 \{\pm 1\}$. In addition, if $\norm{\cA{\pi} - A^*} \leq \kappa
 \norm{A^*}$ holds, then $A$ is said to be $(\delta, \kappa)$-near to $A^*$.
\label{def_closeness}
\end{Definition}

To keep notation simple, in our convergence theorems below, whenever we discuss nearness, we simply replace the transformations $\pi$ and $\sigma$ in the above definition with the identity mapping $\pi(i) = i$ and the positive sign $\sigma(\cdot) = +1$ while keeping in mind that in reality, we are referring to finding one element in the equivalence class of all permutations and sign flips of $A^*$.


Armed with the above concepts, we now posit a generative
model for our observed data. Suppose that the data samples $Y = [y^{(1)},
y^{(2)}, \dots, y^{(p)}]$ are such that each column is generated according to the rule:
\begin{equation}
\label{eq:gen}
y = \PGam(A^*x^*),
\end{equation} 
where $A^*$ is an unknown, ground truth dictionary; $x^*$ and $\Gamma$ are drawn from some distribution $\mathcal{D}$ and $\PGam$ is the sampling operator that keeps entries in $\Gamma$ untouched and zeroes out everything else. We emphasize that $\Gamma$ is independently chosen for each $y^{(i)}$, so more precisely, $y^{(i)} = y^{(i)}_{\Gamma^{(i)}} \in \R^n$. We ignore the superscript to keep the notation simple. We also make the following assumptions:


\begin{Assumption}
  The true dictionary $A^*$ is over-complete with $m \leq Kn$ for some constant $K > 1$, and democratic with parameter $\mu$. All columns of $A^*$ have unit norms.
\end{Assumption}

\begin{Assumption}
\label{assum:sparse}
 The true dictionary $A^*$ has bounded spectral and max ($\ell_\infty$-vector) 
 norms such that $\norm{A^*} \leq
  O(\sqrt{m/n})$ and $\norm{A^*}_{\max} \leq O(1/\sqrt{n})$.
\end{Assumption}

\begin{Assumption}
  The code vector $x^*$ is $k$-sparse random with uniform support $S$. 
  The nonzero entries of $x^*$ are pairwise independent sub-Gaussian with variance 1, and bounded
  below by some known constant $C$.
\end{Assumption}

\begin{Assumption}
Each entry of the sample $A^*x^*$ is independently observed with constant probability
$\rho \in (0,1]$. 
\end{Assumption}

The incoherence and spectral bound are ubiquitous in the dictionary learning
literature~\citep{arora-colt14,arora-colt15}. For the
incomplete setting, we further require the democracy and max-norm
bounds to control the spread of energy of the entries of $A^*$, so that $A^*$ is not ``spiky''. Such
conditions are often encountered in the matrix completion literature~\citep{mc1,mc2}.  
The distributional assumptions on the code vectors $x^*$ are standard
in theoretical dictionary learning~\citep{agarwal14-learning,arora-colt14,gribonval15-sample,arora-colt15}. Finally, we also require the sparsity $k \leq
O^*(\rho\sqrt{n}/\log n)$ throughout the
paper.


\section{A Descent-Style Learning Algorithm}
\label{descent}

We now design and analyze an algorithm for learning the dictionary $A^*$ given incomplete samples of the form \eqref{eq:gen}. Our strategy will be to use a descent-like scheme to construct a sequence of estimates $A$ which successively gets closer to $A^*$ in the sense of $(\delta,\kappa)$-nearness.


Let us first provide some intuition. The natural approach to solve this problem is to perform gradient descent over an appropriate empirical loss of the dictionary parameters. More precisely, we consider the squared loss between observed entries of $Y$ and their estimates (which is the typical loss function used in the incomplete observations setting~\citep{jain13-lowrank}):
\begin{align}
  \begin{split}
    \mathcal{L}(A) &= \frac{1}{2}\sum_{i, j \in \Omega} (Y_{ij} - (AX)_{ij})^2,\label{eq_loss} \\
  \end{split}
\end{align}
where $\Omega$ is the set of locations of observed entries in the
samples $Y$. However, straightforward gradient descent over $A$ is not possible for several reasons: (i) the gradient depends on the finite sample variability of $Y$; (ii) the gradient with respect to $A$ depends on the optimal code vectors of the data samples, $x_i^*$, which are unknown \emph{a priori}; (iii) since we are working in the overcomplete setting, care has to be taken to ensure that the code vectors (\ie, columns of $X$) obey the sparsity model (as specified in Assumption~\ref{assum:sparse}). 


The \emph{neurally-plausible sparse coding} algorithm of~\citet{arora-colt15} provides a crucial insight into the understanding of the loss surface of $\mathcal{L}_A$ in the fully observed setting. Basically, within a small ball around the ground truth $A^*$, the surface is well behaved such that a \emph{noisy} version of $X^*$ is sufficient to construct a good enough approximation to the gradient of $\mathcal{L}$. Moreover, given an estimate within a small ball around $A^*$, a noisy (but good enough) estimate of $X^*$ can be quickly computed using a thresholding operation. 

We extend this understanding to the (much more challenging) setting of
incomplete observations. Specifically, we show the loss surface
in~\eqref{eq_loss} behaves well even with missing data. This enables
us to devise an algorithm similar to that of~\citet{arora-colt15} and obtain a descent property directly related to (the population parameter) $A^*$. The full procedure is detailed as Algorithm~\ref{alg_descent}. 

\begin{algorithm}[!t] 
  \caption{Gradient descent-style algorithm}
  \label{alg_descent}
  \SetAlgoLined
  \textbf{Input:} \\
  Partial samples $Y$ with observed entry set $\Gamma^{(i)}$ \\
  Initial $A^0$ that is $(\delta, 2)$-near to $A^*$ \\
  \For{$s = 0, 1, \dots, T$}{
    \tcc{Encoding step}
    \For{$i = 1, 2, \dots, p$}{ \comment{Encode}
      $x^{(i)} \leftarrow \thres_{C/2}(\frac{1}{\rho}(A^{s})^Ty^{(i)})$
    }
    \tcc{Update step}
    $\ghat^s \leftarrow \frac{1}{p} \sum_{i=1}^p
    (\mathcal{P}_{\Gamma^{(i)}}(A^sx^{(i)}) - y^{(i)})\sgn(x^{(i)})^T$\;\\
    $A^{s+1} \leftarrow  A^s - \eta\ghat^s$ \\ 
  }
  \textbf{Output:} $A \leftarrow A^T$ as a learned dictionary
\end{algorithm}

We now analyze our proposed algorithm. Specifically, we can show that if initialized properly and with proper choice of step size, Algorithm~\ref{alg_descent} exhibits \emph{linear} convergence to a ball of radius $O(\sqrt{k/n})$ around $A^*$
. Formally, we have:

\begin{Theorem}
  \label{main_thm_desc}
  Suppose that the initial estimate $A^0$ is $(\delta, 2)$-near to $A^*$ with $\delta = O^*(1/\log n)$ and the sampling probability satisfies $\rho \geq 1/(k+1)$. If Algorithm~\ref{alg_descent} is given $p = \Omgtilde(mk)$ fresh partial samples at each step and uses learning rate $\eta = \Theta(m/\rho k)$, then
  $$\E[\norm{\cAi^s - \cAi^*}^2] \leq (1- \tau)^s \norm{\cAi^0 -
    \cAi^*}^2 + O(\sqrt{k/n}) $$
  for some $0 < \tau < 1/2$ and $s = 1, 2, \dots, T$. As a corollary, $A^s$ converges geometrically to $A^*$ until column-wise $O(\sqrt{k/n})$ error.
\end{Theorem}
We defer the full proof of Theorem
\ref{main_thm_desc} to Appendix~\ref{app:concen}. 
To understand the working of the algorithm and its correctness, let us consider the setting where we have access to infinitely many samples. This setting is, of course, fictional; however, expectations are easier to analyze than empirical averages, and moreover, this exercise reveals several key elements for proving Theorem \ref{main_thm_desc}. More precisely, we first provide bounds on the expected value of $\ghat^s$, denoted as 
$$g^s \triangleq \E_y[(\PGam(A^sx) - y)\sgn(x)^T],$$ 
to establish the descent property for the infinite sample case. The sample complexity argument emerges when we control the concentration of $\ghat^s$, detailed in Appendix~\ref{app:concen}.
Here, we separately discuss the encoding and update steps in Algorithm~\ref{alg_descent}.

\textbf{Encoding step}. The first main result is to show that the hard-thresholding (or pooling)-based rule for estimating the sparse code vectors is sufficiently accurate. This rule adapts the encoding step of the dictionary learning algorithm proposed in~\citep{arora-colt15}, with an additional scaling factor $1/\rho$. This scaling is necessary to avoid biases arising due to the presence of incomplete information. 

The primary novelty is in our analysis. Specifically, we prove that the estimate of $X$ obtained via the encoding step (even under partial observations) enables a good enough identification of the \emph{support} of the true $X^*$. The key, here, is to leverage the fact that $A^*$ is \emph{democratic} and that $A^s$ is near $A^*$. We call this property \emph{support consistency} and establish it as follows.

\begin{Lemma}
  \label{lm_sign_recovery_x}
  Suppose that $A^s$ is $(\delta, 2)$-near to $A^*$ with $\delta = O^*(1/\log
  n)$. With high probability over $y = \PGam(A^*x^*)$, the estimate $x$ obtained by the encoding step of Algorithm~\ref{alg_descent} has the same sign as the true $x^*$; that is, 
  \begin{equation}
    \label{eq_sign_recovery_x}
    \sgn\bigl(\thres_{C/2}{\bigl(\frac{1}{\rho}(A^{s})^Ty\bigr)}\bigr) = \sgn(x^*),
  \end{equation}
This holds true for incoherence parameter $\mu \leq
\frac{\sqrt{n}}{2k}$, sparsity parameter $k \geq \Omega(\log m)$ and
subsampling probability $\rho \geq 1/(k+1)$.
\end{Lemma}

Lemma~\ref{lm_sign_recovery_x} implies that when the ``mass'' of $A^*$
is spread out across entries, within a small neighborhood of $A^*$ the
estimate $x$ is reliable \emph{even} if $y$ is incompletely observed. This lemma is the main ingredient
for bounding the behavior of the update rule. 

\textbf{Update step}. The support consistency property of the estimated $x$ arising in the encoding step is key to rigorously analyzing the
expected gradient $g^s$. This relatively `simple' encoding enables an explicit form of the update rule, and gives an intuitive reasoning on how the descent
property can be achieved. In fact, we will see that
\begin{align*}
  g_i^s &= \rho p_iq_i(\lambda_i^s\cAi^s - \cAi^*) + o(\rho p_iq_i)
\end{align*}
for $p_i = \E[\abs{x_i^*}|i \in S]$, $q_i = \Prob[i \in S]$ and $\lambda_i^s = \inprod{\cAi}{\cAi^*}$.
Since we assume that the current estimate $A^s$ is (column-wise) sufficiently close to $A^*$, each $\lambda_i^s$ is approximately equal to 1, and hence
$g_i^s \approx \rho p_iq_i(\cAi^s - \cAi^*)$, i.e., the gradient points in the desired direction. Combining this with standard analysis of gradient descent, 
we can prove that the overall algorithm geometrically decreases the error in each step $s$ as long as the learning rate
$\eta$ is properly chosen. Specifically, we get the following theoretical result.

\begin{Theorem}
  \label{thm_desc_inf_sample}
  Suppose that $A^0$ is $(\delta, 2)$-near to $A^*$ with $\delta = O^*(1/\log
  n)$ and the sampling probability satisfies $\rho \geq 1/(k+1)$. Assuming 
  infinitely many partial samples at each step, Algorithm~\ref{alg_descent} geometrically converges to $A^*$ until column-wise error $O(k/\rho n)$. More precisely,
  $$\norm{\cAi^{s+1} - \cAi^*}^2 \leq (1-\tau) \norm{\cAi^s -
    \cAi^*}^2 + O\bigl(k^2/\rho^2 n^2 \bigr) $$
  for some $0 < \tau < 1/2$ and for $s = 1, 2, \dots,
  T$ provided the learning rate obeys $\eta = \Theta(m/\rho k)$. 
\end{Theorem}

We provide the mathematical proof for the form of $g^s$ as well as the
descent in Appendix~\ref{sec:correl}. We also argue that the $(\delta,
2)$-nearness of $A^{s+1}$ and $A^*$ is maintained after each
update. This is studied in Lemma~\ref{lm_nearness} in Appendix~\ref{app:desc}.



\section{An Initialization Algorithm}
\label{init}

In the previous section, we provided an algorithm that (accurately) recovers $A^*$ in an
iterative descent-style approach. In order to establish correctness guarantees, the algorithm requires a coarse
estimate $A^0$ that is $\delta$-close to the ground truth with closeness parameter $\delta = O^*(1/\log n)$. 
This section presents an initialization strategy to obtain such a good starting point for $A^*$.


Again, we begin with some intuition. At a high level, our algorithm mimics the spectral initialization strategy for dictionary learning proposed by~\cite{arora-colt15}. In
essence, the idea is to re-weight the data samples (which are fully observed) appropriately. When this is the case, analyzing the spectral properties of the covariance matrix of the new re-weighted samples
gives us the desired initialization. The re-weighting itself relies upon the computation of pairwise
correlations between the samples with two fixed samples (say, $u$ and $v$) chosen from an
independent \emph{hold-out set}. This strategy is appealing in both from the standpoint of statistical efficiency as well as computational ease.

Unfortunately, a straightforward application of this strategy to our setting of incomplete observations does not work. The major issue, of course, is that pairwise correlation (the inner product) of two high dimensional vectors is highly uninformative if each vector is only partially observed.  We circumvent this issue via the following simple (but key) observation: \emph{provided the underlying dictionary is democratic} and \emph{the representation is sufficiently sparse}, the correlation between a partially observed data sample $y$ with a \emph{fully} observed sample $u$ is indeed proportional to the actual correlation between $y$ and $u$. 
Therefore, assuming that we are given a hold-out set \emph{that is
  fully observed}, an adaptation of the spectral approach of
\citet{arora-colt15} provably succeeds. Moreover, the size of the
hold-out set need not be large; in particular, we need only $O(m~\polylog(n))$
fully-observed samples, as opposed to the $O(mk~\polylog(n))$ samples
required by the analysis of~\citet{arora-colt15}. The parameter $k$ can be as big as $\sqrt{n}$, so in fact we require polynomially fewer fully-observed samples.

In summary: in order to initialize our descent procedure, we assume the availability of a small (but fully observed) hold-out set. In practice, we can imagine expending some amount of effort in the beginning to collect all the entries of a small subset of the available data samples. The availability of such additional information (or ``side-information'') has been made in the literature on matrix completion~\citep{inductive-matrix-completion}.

The full procedure is described in pseudocode form as Algorithm~\ref{alg_init}. Our main theoretical result (Theorem~\ref{main_thm_init}) summarizes its performance.

\begin{algorithm}[!t]
  \caption{Spectral initialization algorithm}
  \label{alg_init}
  \SetAlgoLined
  \textbf{Input:} \\
  $\mathcal{P}_1$: $p_1$
  fully observed samples \\
  $\mathcal{P}_2$: $p_2$ partially
  observed samples \\
  Set $L = \emptyset$ \\
  \While{$|L| < m$}{
    Pick $u$ and $v$ from $\mathcal{P}_1$ at random\; \\
    Construct the weighted covariance matrix $\Mhuv$ using samples $y^{(i)}$ from $\mathcal{P}_2$
    $$\Mhuv \leftarrow \frac{1}{p_2\rho^4}\sum_{i=1}^{p_2}\langle y^{(i)}, u\rangle
    \langle y^{(i)}, v\rangle y^{(i)}(y^{(i)})^T$$
    $\delta_1, \delta_2 \leftarrow $ top singular values\; \\
    \If{$\delta_1 \geq \Omega(k/m)$ and $\delta_2 < O^*(k/m\log n)$}{
      $z \leftarrow $ top singular vector\; \\
      \If{$z$ is not within distance $1/\log n$ of vectors in $L$
        even with sign flip}{
        $L \leftarrow L \cup \{z\}$
      }
   }
 }
 \textbf{Output:} $A^0 \leftarrow \text{Proj}_{\mathcal{B}}(\tilde{A})$ where $\tilde{A}$ is the matrix whose columns in $L$ and $\mathcal{B} = \{A: \norm{A} \leq 2\norm{A^*}\}$
\end{algorithm}

\begin{Theorem}
  \label{main_thm_init}
Suppose that the available training dataset consists of $p_1$ fully observed samples, together with $p_2$ incompletely observed samples according to the observation model \eqref{eq:gen}. Suppose $\mu =
  O^*\bigl(\frac{\sqrt{n}}{k\log^3n}\bigr)$, $\frac{1}{\rho} -1 \leq
  k \leq O^*(\frac{\rho\sqrt{n}}{\log n})$. 
  When $p_1 = \Omgtilde(m)$ and $p_2 =
  \Omgtilde(mk/\rho^4)$, then with high probability, Algorithm~\ref{alg_init} returns an initial estimate $A^0$ whose columns share the same support as $A^*$ and is $(\delta, 2)$-near to $A^*$ with $\delta = O^*(1/\log n)$.
\end{Theorem}

The full proof is provided in Appendix~\ref{app:init}. To provide some intuition about the working of the algorithm and its proof, let us again consider the setting where we have access to infinitely many samples. 
These analyses result in key lemmas, which we will reuse extensively for proving Theorem~\ref{main_thm_init}.

First, consider two \emph{fully observed} data samples $u = A^*\alpha$ and $v = A^*\alpha'$ drawn from the hold-out set. (Here, $A^*, \alpha, \alpha'$ are unknown.) Consider also a partially observed sample $y =
\rAG^*x^*$ under a random subset $\Gamma \subseteq [n]$. Define:
$$\beta = \frac{1}{\rho} \rAG^{*T}u,~\text{and } \beta' = \frac{1}{\rho}
\rAG^{*T}v$$
respectively as (crude) estimates of $\alpha$ and $\alpha'$, simply obtained by applying a (scaled) adjoint of $\rAG$ to $u$ and $v$ respectively. 
It follows from the above definition that:
$$
\beta = \frac{1}{\rho} \rAG^{*T} A^* \alpha,~\text{and}~\langle y, u \rangle = \rho \langle \beta, x^* \rangle .
$$
Our main claim is that since $A^*$ is assumed to satisfy the democracy property, $\frac{1}{\rho}\rAG^{*T} A^*$ resembles the identity, and hence $\beta$ ``looks'' like the true code vector $\alpha$. In particular, we have the following lemma.
\begin{Lemma}
  \label{cl_beta_estimate}
  With high probability over the randomness in $u$ and $\Gamma$, we have: (a)
  $\abs{\beta_i - \alpha_i} \leq \frac{\mu k\log n}{\sqrt{n}} +
  \sqrt{\frac{1-\rho}{\rho n^{1/2}}}$ for each $i = 1, 2, \dots, m$ and
  (b) $\norm{\beta} \leq \frac{\sqrt{k}\log n}{\rho}$.
\end{Lemma}
\proof Denote $U = \supp(\alpha)$ and $W = U \backslash
\{i\}$, then 
\begin{align}
  \label{init_eq:1}
  \abs{\beta_i - \alpha_i} &= \abs[\Big]{\frac{1}{\rho}\AG{i}^{*T}\cA{W}^*\alpha_W +
                             \bigl(\frac{1}{\rho}\inprod{\AG{i}^*}{\cAi^*}-1\bigr)\alpha_i}
  \nonumber \\
                           &\leq \frac{1}{\rho}\abs[\big]{\AG{i}^{*T}\cA{W}^*\alpha_W} + \abs[\Big]{(\frac{1}{\rho}\AG{i}^{*T}\cAi^*-1)\alpha_i} .
\end{align}
We will bound these terms on the right hand side of~\eqref{init_eq:1} using the properties of $A^*$ and $\alpha$. First, we notice that for any $\Gamma \subset [n]$:
\begin{align*}
  \norm{\AG{i}^{*T}\cA{W}^*}^2 &= \sum_{j \in
                                 W}\inprod{\AG{i}^*}{\cAj^*}^2 
                               \leq \frac{\mu^2}{n}\sum_{j \in
  W}\norm{\AG{i}^*}^2\norm{\AG{j}^*}^2 ,
\end{align*}
where we have used the democracy of $A^*$ with respect to $\Gamma$. Moreover, using the Chernoff bound for
$\norm{\AG{i}^*}^2 = \sum_{i=1}^n A_{li}^{*2}\1[l \in \Gamma]$, we have $\norm{\AG{i}^*}^2 \leq \rho + o(\rho)$ \whp\ Hence,
$\norm{\AG{i}^{*T}\cA{W}^*}^2 \leq \rho^2\mu^2k/n$ with high probability.
In addition, $\norm{\alpha_W} \leq \sqrt{k}\log
n$ \whp\ because $\alpha_W$ is $k$-sparse sub-Gaussian. Therefore, the first term in~\eqref{init_eq:1} gives $\frac{1}{\rho}\abs{\AG{i}^{*T}\cA{W}^*\alpha_W} \leq
\frac{\mu k\log n}{\sqrt{n}}$ with high probability.

For the second term in~\eqref{init_eq:1}, consider a random variable $T =
(\frac{1}{\rho}\AG{i}^{*T}\cAi^*- 1)\alpha_i$ over $\Gamma$ and
$\alpha_i$. We first observe for any vector $w \in \R^n$ that:
\begin{align*}
  \E[(w_{\Gamma}^Tw)^2] &= \sum_{i =1}^n\E[w_i^4\1_{i \in \Gamma}] + \sum_{i \neq j}^n\E[w_i^2w_j^2\1_{i,j \in \Gamma}] \\
                        &= \rho(1-\rho)\sum_{i =1}^nw_i^4 + \rho^2.
\end{align*}
Hence, $T$ has mean 0 and variance
$\sigma^2_T = (1-\rho)/\rho\sum_{j=1}^nA_{ji}^4$, which is
bounded by $O(\frac{1-\rho}{\rho n})$ because $\norm{A^*}_{\max} \leq O(1/\sqrt{n})$. By Chebyshev's inequality,
we have $\abs{T} \leq \sqrt{\frac{1-\rho}{\rho n^{1/2}}}$ with failure probability $1/\sqrt{n}$. 
Combining everything, we get 
$$\abs{\beta_i - \alpha_i} \leq
\frac{\mu k\log n}{\sqrt{n}} + \sqrt{\frac{1-\rho}{\rho
    n^{1/2}}},$$ \whp, which is the first part of the claim.

For the second part, we bound $\norm{\beta}$ by expanding it as:
$$\norm{\beta} = \frac{1}{\rho}\norm{\rAG^{*T}\cA{U}^*\alpha_U} \leq
\frac{1}{\rho}\norm{\rAG^*}\norm{\cA{U}^*}\norm{\alpha_U},$$
and again, if we use $\norm{\alpha_U} \leq
\sqrt{k}\log n$ \whp and $\norm{A^*} \leq O(1)$, then $\norm{\beta} \leq \sqrt{k}\log n/\rho$. \qedhere 

We briefly compare the above result with that of~\citet{arora-colt15}. Our upper bounds are more general, and are stated in terms of the incompleteness factor $\rho$. Indeed, our results match the previous bounds when $\rho = 1$. The above lemma suggests the following interesting regime of parameters. Specifically, for $\mu = O^*\bigl(\frac{\sqrt{n}}{k\log^3n}\bigr)$ and $\frac{1}{\rho} -1 \leq  k \leq O^*(\frac{\rho\sqrt{n}}{\log n})$, one can see that $\abs{\beta_i - \alpha_i} \leq O^*(1/\log^2n)$ \whp , which implies that $\beta$ is a good estimate
of $\alpha$ even when a subset of rows in $A^*$ is given. 

In the next lemma, we show that that the pairwise correlation of 
$u$ and any sample $y$ is sufficiently informative for the same re-weighted spectral estimation strategy of~\citet{arora-colt15} to succeed in the incomplete setting.

\begin{Lemma}
  \label{lm_Muv}
  Suppose that $u, v$ are a pair of fully observed samples and $y$ is an incomplete sample
  independent of $u, v$. The weighted covariance matrix $M_{u, v}$ has the form:
  \begin{align*}
    M_{u, v} &\triangleq \frac{1}{\rho^4}\E_y[\inprod{y}{u}\inprod{y}{v}yy^T] \\
             &= \sum_{i \in U \cap V}
               q_ic_i\beta_i\beta'_i\cAi^*\cAi^{*T} + O^*(k/m\log n) ,
  \end{align*}
  where $c_i = \E[x_i^{*4}|i\in S]$ and $q_i = \Prob[i \in S]$. 
\end{Lemma}
The complete proof is relegated to Appendix~\ref{app:init}. We will instead
discuss some implications of this Lemma. Recall that $c_i$ is a constant with $0<c<1$ and $q_i = \Theta(k/m)$. 

Suppose, for a moment, that the sparse representations of $u$ and $v$ share exactly one common dictionary element, say $\cAi^*$ (i.e., if $U = \supp(u)$ and $V = \supp(v)$ then $U \cap V = \{i\}$.)
The first term, $q_ic_i\beta_i\beta'_i\cAi^*\cAi^{*T}$, has norm
$\abs{q_ic_i\beta_i\beta'_i}$. From Claim~\ref{cl_beta_estimate},
$\abs{\beta_i} \geq \abs{\alpha_i} - \abs{\beta_i - \alpha_i}\geq C -
o(1)$. Therefore, $q_ic_i\beta_i\beta'_i\cAi^*\cAi^{*T}$ has norm at
least $\Omega(k/m)$ whereas the perturbation terms are at most
$O^*(k/m\log n)$. According to Wedin's theorem, we conclude that the top singular
vector of $M_{u,v}$ must be $O^*(k/m\log n)/\Omega(k/m) = O^*(1/\log
n)$ -close to $\cAi^*$. This gives us (an estimate of) one of the columns of $A^*$. 

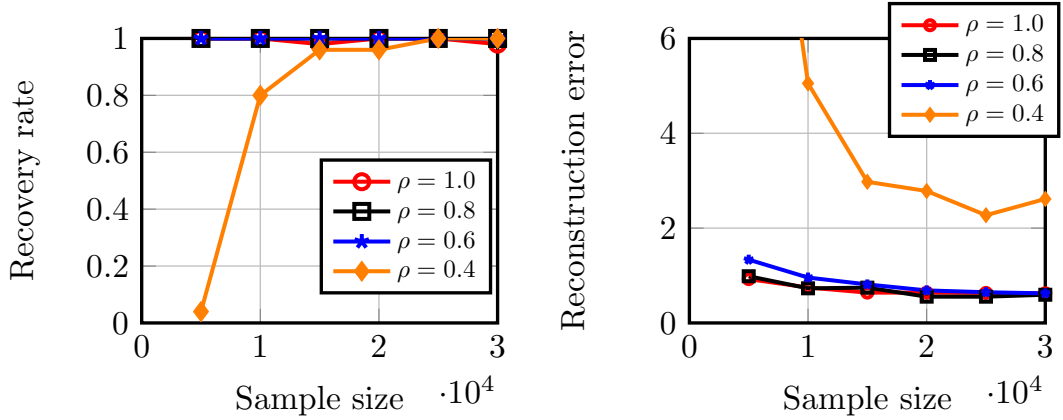
\begin{figure}[!t]
\begin{center}
\resizebox{0.9\columnwidth}{!}{
\begin{tabular}{ccc}
\begin{tikzpicture}[scale=1.2]
\begin{axis}[
		width=3.75cm,
		height=3cm,
		scale only axis,
		xmin=0, xmax=30000,
		xlabel = {Sample size},
		xmajorgrids,
		ymin=0, ymax=1,
		ylabel={Recovery rate},
		ymajorgrids,
		line width=1.0pt,
		mark size=1.5pt,
		legend style={nodes={scale=0.75, transform shape},at={(0.50,.1)},anchor=south west,draw=black,fill=white,align=left}
		]
\addplot  [color=red,
		solid, 
		very thick,
		mark=o,
		mark options={solid,scale=1.5},
		]
		table [x index = 0,y index=1]{results/prob_success_sparfa_mc50_k6.txt};
\addlegendentry{$\rho =1.0$}
\addplot [color=black,
		solid, 
		very thick,
		mark=square,
		mark options={solid,scale=1.5},
		]
		table [x index = 0,y index=2]{results/prob_success_sparfa_mc50_k6.txt};
\addlegendentry{$\rho = 0.8$}
\addplot  [color=blue,
		solid, 
		very thick,
		mark=star,
		mark options={solid,scale=1.5},
		]
		table [x index = 0,y index=3]{results/prob_success_sparfa_mc50_k6.txt};
\addlegendentry{$\rho =0.6$}
\addplot [color=orange,
		solid, 
		very thick,
		mark=diamond*,
		mark options={solid,scale=1.5},
		]
		table [x index = 0,y index=4]{results/prob_success_sparfa_mc50_k6.txt};
\addlegendentry{$\rho = 0.4$}
\end{axis}
\end{tikzpicture}
&
\begin{tikzpicture}[scale=1.2]
\begin{axis}[
		width=3.75cm,
		height=3cm,
		scale only axis,
		xmin=0, xmax=30000,
		xlabel = {Sample size},
		xmajorgrids,
		ymin= 0, ymax=6,
		ylabel={Reconstruction error},
		ymajorgrids,
		line width=1.0pt,
		mark size=1.0pt,
		legend style={nodes={scale=0.75, transform shape},at={(0.8,.65)},anchor=south,draw=black,fill=white,align=left}           
		]
\addplot  [color=red,
		solid, 
		very thick,
		mark=o,
		mark options={solid,scale=1.5},
		]
		table [x index = 0,y index=1]{results/error_sparfa_mc50_k6.txt};
\addlegendentry{$\rho=1.0$}
\addplot [color=black,
		solid, 
		very thick,
		mark=square,
		mark options={solid,scale=1.5},
		]
		table [x index = 0,y index=2]{results/error_sparfa_mc50_k6.txt};
\addlegendentry{$\rho=0.8$}
\addplot  [color=blue,
		solid, 
		very thick,
		mark=star,
		mark options={solid,scale=1.5},
		]
		table [x index = 0,y index=3]{results/error_sparfa_mc50_k6.txt};
\addlegendentry{$\rho =0.6$}
\addplot [color=orange,
		solid, 
		very thick,
		mark=diamond*,
		mark options={solid,scale=1.5},
		]
		table [x index = 0,y index=4]{results/error_sparfa_mc50_k6.txt};
\addlegendentry{$\rho = 0.4$}

\end{axis}
\end{tikzpicture}
\end{tabular}
}
\caption{\small\sl (top) The performance of our approach on two metrics (recovery rate and reconstruction error) in sample size and subsampling probability. \label{fig_simulation_results}}
\end{center}
\end{figure}

The question remains when and how whether we can \emph{a priori} certify whether $u, v$ share a unique dictionary atom among their sparse representations. Fortunately, the following Lemma provides a simple test for this via examining the decay of the singular vectors of the cross-covariance matrix $M_{u,v}$. The proof follows directly from that of Lemma 37 in~\citep{arora-colt15}.
\begin{Lemma}
  \label{lm_condition_uv_share_unique_supp}
  When the top singular value of $M_{u,v}$ is at least $\Omega(k/m)$ and the second largest one is at most $O^*(k/m\log n)$, then $u$ and $v$ share a unique dictionary element with high probability. 
\end{Lemma}

The above discussion isolates one of the columns of $A^*$. We can repeat this procedure several times by randomly choosing pairs of samples $u$ and $v$ from the hold-out set.
 Using the result of~\citet{arora-colt15}, if $|\mathcal{P}_1|$ is $p_1 = \Otilde(m)$, then we can estimate all the $m$ dictionary atoms. Overall, the sample complexity of Algorithm~\ref{alg_init} is dominated by $p_2 = \Otilde(mk/\rho^4)$. 


\section{Experiments}
\label{exp}

We corroborate our theory by demonstrating some representative numerical benefits of our proposed algorithms. 

We generate a synthetic dataset based on the generative model described in Section~\ref{model}. The ground truth dictionary $A^*$ is of size $256\times 256$ with independent standard Gaussian entries. We normalize columns of $A^*$ to be unit norm. Then, we generate $6$-sparse code vectors $x^*$ with support drawn
uniformly at random. Entries in the support are sampled from $\pm 1$ with equal probability. We generate all full samples, and isolate 5000 samples as {``side information''} for the initialization step. The remaining are then subsampled with different parameters $\rho$.

We set the number of iterations to $T = 3000$ in the initialization
procedure and the number of descent steps $T= 50$ for the descent
scheme. Besides, we slightly modify the thresholding operator in the
encoding step of Algorithm~\ref{alg_descent}. We use another operator
that keeps $k$ largest entries of the input untouched and sets
everything else to zero due to its stability. For each Monte Carlo
trial, we uniformly draw $p$ partial samples. The task, for our
algorithm, is to learn $A^*$. A Matlab implementation of all the
algorithms is available online\footnote{{\small \url{https://github.com/thanh-isu/sparse-coding-for-missing-data}}}.

We evaluate our algorithm on two metrics against $p$ and $\rho$: (i) recovery rate, i.e., the fraction of trials in which each algorithm successfully recovers the ground truth $A^*$; and
 (ii) reconstruction error. All the metrics are averaged over 50 Monte Carlo simulations. ``Successful recovery'' is defined according to a threshold $\tau = 6$ on the Frobenius norm of the difference between the estimate $\widehat{A}$ and the ground truth $A^*$. (Since we can only estimate $\widehat{A}$ modulo a permutation and sign flip, the optimal column and sign matching is computed using the Hungarian algorithm.)

Figure \ref{fig_simulation_results} shows our experimental results. Here, sample size refers to the number of incomplete samples. Our algorithms are able to recover the dictionary for $\rho = 0.6, 0.8, 1.0$. For $\rho=0.4$, we can observe a ``phase transition'' in sample complexity of successful recovery around $p = 10,000$ samples. 
We intend to explore more thorough numerical experiments on realistic (and larger) datasets as part of future work.


\bibliography{icml18}

\begin{thebibliography}{31}
\providecommand{\natexlab}[1]{#1}
\providecommand{\url}[1]{\texttt{#1}}
\expandafter\ifx\csname urlstyle\endcsname\relax
  \providecommand{\doi}[1]{doi: #1}\else
  \providecommand{\doi}{doi: \begingroup \urlstyle{rm}\Url}\fi

\bibitem[Elad and Aharon(2006)]{elad06-denoising}
Michael Elad and Michal Aharon.
\newblock Image denoising via sparse and redundant representations over learned
  dictionaries.
\newblock \emph{IEEE Transactions on Image processing}, 15\penalty0
  (12):\penalty0 3736--3745, 2006.

\bibitem[Aharon et~al.(2006)Aharon, Elad, and Bruckstein]{aharon06-ksvd}
Michal Aharon, Michael Elad, and Alfred Bruckstein.
\newblock $k$-svd: An algorithm for designing overcomplete dictionaries for
  sparse representation.
\newblock \emph{IEEE Transactions on Signal Processing}, 54\penalty0
  (11):\penalty0 4311--4322, 2006.

\bibitem[Olshausen and Field(1997)]{olshausen97-sc}
Bruno~A Olshausen and David~J Field.
\newblock Sparse coding with an overcomplete basis set: A strategy employed by
  v1?
\newblock \emph{Vision research}, 37\penalty0 (23):\penalty0 3311--3325, 1997.

\bibitem[Candes and Tao(2005)]{candes05-decoding}
Emmanuel~J Candes and Terence Tao.
\newblock Decoding by linear programming.
\newblock \emph{IEEE Transactions on Information Theory}, 51\penalty0
  (12):\penalty0 4203--4215, 2005.

\bibitem[Rubinstein et~al.(2010)Rubinstein, Bruckstein, and Elad]{elad2}
Ron Rubinstein, Alfred~M Bruckstein, and Michael Elad.
\newblock Dictionaries for sparse representation modeling.
\newblock \emph{Proceedings of the IEEE}, 98\penalty0 (6):\penalty0 1045--1057,
  2010.

\bibitem[Gregor and LeCun(2010)]{lista}
Karol Gregor and Yann LeCun.
\newblock Learning fast approximations of sparse coding.
\newblock In \emph{International Conference on Machine Learning (ICML)}, pages
  399--406, 2010.

\bibitem[Boureau et~al.(2010)Boureau, Bach, LeCun, and
  Ponce]{boureau2010learning}
Y-Lan Boureau, Francis Bach, Yann LeCun, and Jean Ponce.
\newblock Learning mid-level features for recognition.
\newblock In \emph{IEEE Conference on Computer Vision and Pattern Recognition
  (CVPR)}, pages 2559--2566. IEEE, 2010.

\bibitem[Spielman et~al.(2012)Spielman, Wang, and Wright]{spielman12-exact}
Daniel~A Spielman, Huan Wang, and John Wright.
\newblock Exact recovery of sparsely-used dictionaries.
\newblock In \emph{Conference on Learning Theory}, pages 37--1, 2012.

\bibitem[Agarwal et~al.(2013)Agarwal, Anandkumar, and
  Netrapalli]{agarwal13-exact}
Alekh Agarwal, Animashree Anandkumar, and Praneeth Netrapalli.
\newblock Exact recovery of sparsely used overcomplete dictionaries.
\newblock \emph{IEEE Transactions on Information Theory}, 1050:\penalty0 8,
  2013.

\bibitem[Agarwal et~al.(2014)Agarwal, Anandkumar, Jain, Netrapalli, and
  Tandon]{agarwal14-learning}
Alekh Agarwal, Animashree Anandkumar, Prateek Jain, Praneeth Netrapalli, and
  Rashish Tandon.
\newblock Learning sparsely used overcomplete dictionaries.
\newblock In \emph{Conference on Learning Theory}, pages 123--137, 2014.

\bibitem[Arora et~al.(2014)Arora, Ge, and Moitra]{arora-colt14}
Sanjeev Arora, Rong Ge, and Ankur Moitra.
\newblock New algorithms for learning incoherent and overcomplete dictionaries.
\newblock In \emph{Conference on Learning Theory}, pages 779--806, 2014.

\bibitem[Arora et~al.(2015)Arora, Ge, Ma, and Moitra]{arora-colt15}
Sanjeev Arora, Rong Ge, Tengyu Ma, and Ankur Moitra.
\newblock Simple, efficient, and neural algorithms for sparse coding.
\newblock In \emph{Conference on Learning Theory}, pages 113--149, 2015.

\bibitem[Sun et~al.(2015)Sun, Qu, and Wright]{sun15-complete}
Ju~Sun, Qing Qu, and John Wright.
\newblock Complete dictionary recovery using nonconvex optimization.
\newblock In \emph{International Conference on Machine Learning (ICML)}, pages
  2351--2360, 2015.

\bibitem[Chatterji and Bartlett(2017)]{bartlett-nips17}
Niladri Chatterji and Peter Bartlett.
\newblock {A}lternating minimization for dictionary learning with random
  initialization.
\newblock 2017.
\newblock arXiv:1711.03634v1.

\bibitem[Xing et~al.(2012)Xing, Zhou, Castrodad, Sapiro, and
  Carin]{xing-siam12}
Zhengming Xing, Mingyuan Zhou, Alexey Castrodad, Guillermo Sapiro, and Lawrence
  Carin.
\newblock Dictionary learning for noisy and incomplete hyperspectral images.
\newblock \emph{SIAM Journal on Imaging Sciences}, 5\penalty0 (1):\penalty0
  33--56, 2012.

\bibitem[Naumova and Schnass(2017{\natexlab{a}})]{naumova-17}
Valeriya Naumova and Karin Schnass.
\newblock Dictionary learning from incomplete data.
\newblock \emph{arXiv preprint arXiv:1701.03655}, 2017{\natexlab{a}}.

\bibitem[Soni et~al.(2016)Soni, Jain, Haupt, and Gonella]{soni-sfm16}
Akshay Soni, Swayambhoo Jain, Jarvis Haupt, and Stefano Gonella.
\newblock Noisy matrix completion under sparse factor models.
\newblock \emph{IEEE Transactions on Information Theory}, 62\penalty0
  (6):\penalty0 3636--3661, June 2016.

\bibitem[Loh and Wainwright(2011)]{loh-nips11}
Po-Ling Loh and Martin~J Wainwright.
\newblock High-dimensional regression with noisy and missing data: Provable
  guarantees with non-convexity.
\newblock In \emph{Neural Information Processing Systems}, pages 2726--2734,
  2011.

\bibitem[Yuan and Zhang(2013)]{sparsepca}
Xiao-Tong Yuan and Tong Zhang.
\newblock Truncated power method for sparse eigenvalue problems.
\newblock \emph{Journal of Machine Learning Research}, 14\penalty0
  (Apr):\penalty0 899--925, 2013.

\bibitem[Cai et~al.(2016)Cai, Li, Ma, et~al.]{wirtinger}
T~Tony Cai, Xiaodong Li, Zongming Ma, et~al.
\newblock Optimal rates of convergence for noisy sparse phase retrieval via
  thresholded wirtinger flow.
\newblock \emph{The Annals of Statistics}, 44\penalty0 (5):\penalty0
  2221--2251, 2016.

\bibitem[Tu et~al.(2016)Tu, Boczar, Simchowitz, Soltanolkotabi, and
  Recht]{procrustes}
Stephen Tu, Ross Boczar, Max Simchowitz, Mahdi Soltanolkotabi, and Ben Recht.
\newblock Low-rank solutions of linear matrix equations via procrustes flow.
\newblock In \emph{International Conference on Machine Learning}, pages
  964--973, 2016.

\bibitem[Jain et~al.(2013)Jain, Netrapalli, and Sanghavi]{jain13-lowrank}
Prateek Jain, Praneeth Netrapalli, and Sujay Sanghavi.
\newblock Low-rank matrix completion using alternating minimization.
\newblock In \emph{ACM Symposium on Theory of Computing}, pages 665--674. ACM,
  2013.

\bibitem[Davenport et~al.(2009)Davenport, Laska, Boufounos, and
  Baraniuk]{davenport-democ09}
Mark~A Davenport, Jason~N Laska, Petros~T Boufounos, and Richard~G Baraniuk.
\newblock A simple proof that random matrices are democratic.
\newblock \emph{arXiv preprint arXiv:0911.0736}, 2009.

\bibitem[Naumova and Schnass(2017{\natexlab{b}})]{naumova-schnass17}
V.~Naumova and K.~Schnass.
\newblock Dictionary learning from incomplete data for efficient image
  restoration.
\newblock In \emph{European Signal Processing Conference (EUSIPCO)}, pages
  1425--1429, Aug 2017{\natexlab{b}}.

\bibitem[Cand{\`e}s and Recht(2009)]{mc1}
Emmanuel~J Cand{\`e}s and Benjamin Recht.
\newblock Exact matrix completion via convex optimization.
\newblock \emph{Foundations of Computational mathematics}, 9\penalty0
  (6):\penalty0 717, 2009.

\bibitem[Keshavan et~al.(2010)Keshavan, Montanari, and Oh]{mc2}
Raghunandan~H Keshavan, Andrea Montanari, and Sewoong Oh.
\newblock Matrix completion from a few entries.
\newblock \emph{IEEE Transactions on Information Theory}, 56\penalty0
  (6):\penalty0 2980--2998, 2010.

\bibitem[Eriksson et~al.(2012)Eriksson, Balzano, and Nowak]{highrank}
Brian Eriksson, Laura Balzano, and Robert Nowak.
\newblock High-rank matrix completion.
\newblock In \emph{Artificial Intelligence and Statistics}, pages 373--381,
  2012.

\bibitem[Lan et~al.(2014)Lan, Waters, Studer, and Baraniuk]{sparfa}
Andrew~S Lan, Andrew~E Waters, Christoph Studer, and Richard~G Baraniuk.
\newblock Sparse factor analysis for learning and content analytics.
\newblock \emph{Journal of Machine Learning Research}, 15\penalty0
  (1):\penalty0 1959--2008, 2014.

\bibitem[Gribonval et~al.(2015)Gribonval, Jenatton, Bach, Kleinsteuber, and
  Seibert]{gribonval15-sample}
R{\'e}mi Gribonval, Rodolphe Jenatton, Francis Bach, Martin Kleinsteuber, and
  Matthias Seibert.
\newblock Sample complexity of dictionary learning and other matrix
  factorizations.
\newblock \emph{IEEE Transactions on Information Theory}, 61\penalty0
  (6):\penalty0 3469--3486, 2015.

\bibitem[Natarajan and Dhillon(2014)]{inductive-matrix-completion}
Nagarajan Natarajan and Inderjit~S Dhillon.
\newblock Inductive matrix completion for predicting gene--disease
  associations.
\newblock \emph{Bioinformatics}, 30\penalty0 (12):\penalty0 i60--i68, 2014.

\bibitem[Nguyen et~al.(2018)Nguyen, Wong, and Hegde]{nguyen-aaai18}
Thanh~V. Nguyen, Raymond K.~W. Wong, and Chinmay Hegde.
\newblock A provable approach for double-sparse coding.
\newblock In \emph{Proc. Conf. American Assoc. Artificial Intelligence (AAAI)},
  Feb. 2018.

\end{thebibliography}
\bibliographystyle{unsrtnat}

\appendix

\section{Analysis of Algorithm~\ref{alg_descent}}
\label{app:desc}


\subsection{Analysis of coding step}
\label{sec:coding-rule}

\proof[Proof of Lemma~\ref{lm_sign_recovery_x}] 
Denote $S = \supp(x^*)$ and skip the superscript $s$ on $A^s$ for
simplicity of notation. We will argue that \whp\ $S = \{i \in [m] :
\frac{1}{\rho}\abs{\inprod{\cAi}{y}} \geq C/2\}$ and $\sgn(\inprod{\cAi^s}{y}) =
\sgn(x_i^*)$ for every $i \in S$.

First, we write the element-wise estimate before thresholding in the encoding step as follows:
\begin{equation}
  \label{enc_eq:1}
  \frac{1}{\rho}\inprod{\cAi}{y} = \frac{1}{\rho}\inprod{\cAi}{\PGam(A^*x^*)} = \frac{1}{\rho}\inprod{\cAi}{ \rA{\Gamma}^*x^*} = \frac{1}{\rho}\inprod{\cAi}{\AG{i}^*}x_i^* + \frac{1}{\rho}\sum_{j \neq i}\inprod{\cAi}{\AG{j}^*}x_j^*.
\end{equation}
We expect that $Z_i = (1/\rho)\sum_{S\backslash
  \{i\}}\inprod{\cAi}{\AG{j}^*}x_j^*$ is negligible based on the
closeness of $\cAi$ and $\cAi^*$ and the democracy of $A^*$. More
precisely, we want to upper bound it by $C/4$ with high
probability. Here, $C$ is the lower bound of the nonzero coefficients in
$x^*$. In fact, since $\Gamma$ and $x_j^*$ are independent, $Z_i$ is a 
sub-Gaussian random variable with variance
\begin{equation}
  \label{enc_eq:2}
  \sigma^2_{Z_i} =  \frac{1}{\rho^2}\E[\sum_{j \in S\backslash
  \{i\}}\inprod{\cAi}{\AG{j}^*}^2] = \sum_{j \in S\backslash
  \{i\}}\inprod{\cAi}{\cAj^*}^2 + \frac{1-\rho}{\rho}\sum_{j \in S\backslash
  \{i\}, l \in [n]}A_{li}^{2}A_{lj}^{*2}.
\end{equation}
The second term in~\eqref{enc_eq:2} can be bounded by using the facts that $\norm{A^*}_{\max} \leq
O(1/\sqrt{n})$ and $\norm{\cAi} = 1$. 
Specifically,
\begin{align*}
  \sum_{j \in S\backslash \{i\}, l \in [n]}A_{li}^{2}A_{lj}^{*2} ~\leq
  \sum_{j \in S\backslash \{i\}, l \in [n]}O(1/n)A_{li}^{2} \leq
  O(k/n)\norm{\cAi}^2 = O(k/n),
\end{align*}
 Moreover, since 
 $k \leq \rho\sqrt{n}/\log n$, the
 second term in~\eqref{enc_eq:2} is
bounded by $O((1-\rho)/\sqrt{n}\log n) = o(C)$. 

We bound the first term in~\eqref{enc_eq:2} by using the incoherence and
closeness. For each $j \in S\backslash \{i\}$, we have 
$$\inprod{\cAi}{\cAj^*}^2 \leq 2\bigl(\inprod{\cAi^*}{\cAj^*}^2 + \inprod{\cAi - \cAi^*}{\cAj^*}^2\bigr) \leq 2\mu^2/n + 2\inprod{\cAi - \cAi^*}{\cAj^*}^2,$$
since $\abs{\inprod{\cAi^*}{\cAj^*}} \leq \mu/\sqrt{n}$ due to the
$\mu$-incoherence of $A^*$. Now, we combine the term across $j$ and get a
matrix form to leverage the spectral norm bound. 
In particular,
$$\sum_{S\backslash
  \{i\}}\inprod{\cAi}{\cAj^*}^2 \leq  2\mu^2k/n + 2\norm{\cAS^{*T}(\cAi - \cAi^*)}_F^2 \leq 2\mu^2k/n + 2\norm{\cAS^*}^2\norm{\cAi - \cAi^*}^2 \leq O(1/\log n),$$
where we have used $m = O(n)$ and $\norm{\cAS^{*}} \leq
O(1)$. Also, we made use of the condition that $\mu \leq
\frac{\sqrt{n}}{2k}$ and $k=\Omega(\log n)$. Putting these together, we get
$\sigma^2_{Z_i} \leq O(1/\log n)$. By an application of Bernstein's inequality, we get that $\abs{Z_i} \leq C/4$ \whp\

We now argue that 
$(1/\rho)\inprod{\cAi}{y}$ is small when $i
\notin S$ and big otherwise. Clearly, when $i \notin S$,
$(1/\rho)\inprod{\cAi}{y} = Z_i$ is
less than $C/4$ in magnitude \whp\ 
On the contrary, 
when $i \in S$, then $\abs{x_i^*} \geq C$, and using the Chernoff bound for
$\inprod{\cAi}{\AG{i}^*} = \sum_{l=1}^n A_{li}A_{li}^{*}\1[l \in
\Gamma]$, we see that
$$(1/\rho)\inprod{\cAi}{\AG{i}^*} \geq \inprod{\cAi}{\cAi^*} - o(1)
\geq 1 - o(1)$$ 
\whp\ because $\inprod{\cAi}{\cAi^*} \geq  1 -
\delta^2/2$. Hence, $\abs{(1/\rho)\inprod{\cAi}{y}} \geq C/2$ holds
with high probability.

Finally, we take the union bound over all $i = 1, 2, \dots, m$ to finish the proof. \qedhere

\subsection{Analysis of the update $g^s$ (in expectation)}
\label{sec:correl}

Lemma~\ref{lm_sign_recovery_x} is the key to analyzing the approximate gradient
update term 
$$g^s = \E_y[(\PGam(A^sx) - y)\sgn(x)^T].$$ 
This section presents a rigorous analysis of $g^s$, and is a key step towards achieving the descent property stated in
Theorem~\ref{thm_desc_inf_sample}. In essence, we make use of the
distributions of $x^*$, together with its estimate, $x$ to simplify the expectation in $g^s$. The result is the following:
\begin{Lemma}
  \label{lm_gs_form}
  The column-wise expected value $g^s$ of the update rule is of the form
  \begin{align*}
  g_i^s &= \rho p_iq_i(\lambda_i^s\cAi^s \begin{aligned}[t] &- \cAi^*) + \rho p_i\cA{-i}^s\diag(q_{ij})\cA{-i}^{sT}\cAi^* + (1 - \rho)p_iq_i\diag(\cAi^*\circ \cAi^s)\cAi^s \\
        &+ (1-\rho)\sum_{j\neq i}p_iq_{ij}\diag(\cAj^*\circ \cAj^s)\cAj^s ~\pm \gamma,
        \end{aligned}
  \end{align*}
  where $p_i = \E[x_i\sgn(x_i^*)|i\in S]$, $q_i = \Prob[i \in S]$ and $q_{ij} = \Prob[i,j \in S]$. Additionally, $\lambda_i^s = \inprod{\cAi^s}{\cAi^*}$ and $\cA{-i}^s$ denotes $A^s$ with its $i^{\textrm{th}}$ column removed.
In particular, if $A^s$ is $(\delta, 2)$-near to $A^*$ for $\delta =
O^*(1/\log n)$, then all the additive terms in $g_i^s$, except the first term,
have norm of order $o(\rho p_iq_i)$.
\end{Lemma}

\proof 
%
For notational simplicity, we skip the superscript $s$ on $A^s$ and $g^s$. Recall from Lemma \ref{lm_sign_recovery_x} that the sign of $x^*$ is recovered \whp\ from the encoding step. Then under the event that $\supp(x) = \supp(x^*) \equiv S$, we can write $Ax = A_Sx_S = \frac{1}{\rho} A_SA_S^Ty$. Let us consider the $i^{\textrm{th}}$ column of $g$, $g_i$, given by:
\begin{align*}
  g_i &= \E[(\frac{1}{\rho}\AG{S}A_S^T - I) \, y \, \sgn(x_i)] ~\pm \gamma \\
   &=  \E[(\frac{1}{\rho}\AG{S}A_S^T - I) \, y \, \sgn(x_i^*) ] ~\pm \gamma \\
  &= \E[(\frac{1}{\rho}\AG{S}A_S^T - I)\rA{\Gamma}^*x^*\sgn(x_i^*)] ~\pm \gamma \\
  &= \E[\bigl (\frac{1}{\rho}\sum_{j \in S}\AG{j}\cAj^T - I \bigr)\AG{i}^*x_i^*\sgn(x_i^*)] ~\pm \gamma.
\end{align*}
Here, we make use of the fact that nonzero entries are conditionally independent given the support and have zero mean; therefore $\E[x_j^*\sgn(x_i^*)|S] = 0$ for all $j \neq i$. In the
expression, $\gamma$ denotes any vector whose norm is sufficiently
small because of the sign consistency and bounded $(\PGam(A^sx) -
y)\sgn(x)^T$ (see Claim~\ref{cl_norm_bound_ghi} in Appendix~\ref{app:concen}). 

We continue simplifying the form of $g_i$ by denoting $p_i = \E[x_i^*\sgn(x_i^*)|i \in S]$, $q_i = \Prob[i \in S]$ and $q_{ij} = \Prob[i,j \in S]$. 
Then,
\begin{align*}
  g_i &=  \E_{\Gamma}[\frac{1}{\rho}\sum_{j=1}^mp_jq_{ij}\AG{j}\cAj^T\AG{i}^* - p_iq_i\AG{i}^*] \pm \gamma \\
  &= \frac{1}{\rho}\sum_{j=1}^mp_iq_{ij}\E_{\Gamma}[\AG{j}\AG{i}^{*T}]\cAj -\rho p_iq_i\cAi^* \pm \gamma.
\end{align*}
In the final step, we calculate $\E_{\Gamma}[\AG{j}\AG{i}^{*T}]$ over the random $\Gamma$. One can easily show that
\begin{align*}
  \E_{\Gamma}[\AG{j}\AG{i}^{*T}] = \rho^2 \cAj\cAi^{*T} + \rho(1-\rho) \diag(\cAi^*\circ \cAj),
\end{align*}
where we use $\diag(v)$ to denote a diagonal matrix with entries in $v$ and
$\circ$ to denote the element-wise Hadamard product.
As a result, $g_i$ is expressed as follows: 
\begin{align}
  \label{desc_eq:1}
  g_i &= \rho\sum_{j=1}^mp_jq_{ij}\cAj\cAi^{*T}\cAj +
        (1-\rho)\sum_{j=1}^mp_iq_{ij}\diag(\cAi^*\circ \cAj)\cAj -\rho
        p_iq_i\cAi^* \pm \gamma \nonumber \\
      &= \rho p_iq_i(\lambda_i\cAi \begin{aligned}[t] &- \cAi^*) + \rho p_i\cA{-i}\diag(q_{ij})\cA{-i}^T\cAi^* + (1 - \rho)p_iq_i\diag(\cAi^*\circ \cAi)\cAi \\
        &+ (1-\rho)\sum_{j\neq i}p_iq_{ij}\diag(\cAi^*\circ \cAj)\cAj ~\pm \gamma,
        \end{aligned}
\end{align}
where $\lambda_i = \inprod{\cAi}{\cAi^*}$. Furthermore, $\cA{-i}^T$ denotes the matrix $A$ whose $i^\textrm{th}$ column is removed, and $\diag(q_{ij})$ 
denotes the diagonal matrix of $(q_{i1},
q_{i2}\dots, q_{im})^T$ without entry $q_{ii}=q_i$.

We will prove that $\rho p_iq_i(\lambda_i\cAi - \cAi^*)$ is the
dominant term in~\eqref{desc_eq:1}. In the special case when $\rho =
1$, $g_i$ is well studied in~\citep{arora-colt15}. Here we follow the
same strategy and give upper bounds for the remaining terms. 
First, from the nearness we have $\norm{A} \leq \norm{A-A^*} +
\norm{A^*} \leq O(\sqrt{m/n})$, and also $\norm{\cAi^*} = 1$; hence
\begin{align}
  \label{desc_ineq:1} \norm{\rho p_i\cA{-i}\diag(q_{ij})\cA{-i}^T\cAi^*} &\leq (\rho p_i\max_{j\neq i}q_{ij})\norm{A}^2 \\
  &\leq O(\rho p_iq_i\max_{j\neq i}q_{ij}/q_i) = o(\rho p_iq_i),
\end{align}
for $q_{ij} = \Theta(k^2/m^2)$ and $q_i = \Theta(k/m)$. The remaining
terms can be bounded using the max norm constraint and the closeness of $A$ and $A^*$. More precisely, 
\begin{align}
  \label{desc_ineq:2}
  \norm{\diag(\cAi^*\circ \cAi)\cAi} &\leq \norm{\diag(\cAi^*\circ \cAi)} \\
  &\leq \norm{\diag(\cAi^*\circ \cAi^*)} +
           \norm{\diag(\cAi^*\circ (\cAi - \cAi^*))} \nonumber \\
  &\leq O(1/n) + O(\delta/\sqrt{n}) \nonumber \\
  & \leq O(\delta/\sqrt{n}), \nonumber
\end{align}
since $\norm{A}_{\max} \leq O(1/\sqrt{n})$ and $\norm{\cAi - \cAi^*}
\leq \delta$. Since $(1 -\rho)/\rho \leq k$ and $k \leq
O^*(\rho\sqrt{n}/\log n)$, then $$\norm{(1-\rho)p_iq_i\diag(\cAi^*\circ \cAi)\cAi} \leq
O(\rho p_iq_ik\delta/\sqrt{n}) = o(\rho p_iq_i).$$
Similarly, we have
\begin{align}
  \label{desc_ineq:3}
  \norm[\Big]{\sum_{j\neq i}q_{ij}\diag(\cAi^*\circ \cAj)\cAj}^2 &= \sum_{l =1}^n\bigl(\sum_{j\neq i}q_{ij}A_{li}^{*}A_{lj}^2\bigr)^2\nonumber\\
                                                                 &\leq \sum_{l =1}^n(\max_{j \neq i}q_{ij}\norm{A}_{\max})^2(\sum_{j \neq i}A_{lj}^2)^2 \nonumber \\
                                                                 &\leq (\max_{j \neq i}q_{ij}\norm{A}_{\max})^2\sum_{l =1}^n\norm{\rAl}^4
\end{align}
%
%
Moreover $\norm{\rAl} \leq \norm{A} \leq O(1)$, $\norm{A_{\max}} \leq O(1/\sqrt{n})$ and $k \leq O^*(\rho\sqrt{n}/\log n)$, then
\begin{align}
  \label{ineq:4}
  \norm[\Big]{(1-\rho)\sum_{j\neq i}q_{ij}\diag(\cAj^*\circ \cAj)\cAj} &\leq O\Bigl(\rho p_iq_i\frac{1-\rho}{\rho}\max_{j \neq i}q_{ij}/q_i\Bigr) \\
  &= O\Bigl(\rho p_iq_i\frac{(1-\rho)k}{m\rho}\Bigr) = o(\rho p_iq_i)
\end{align}
\qedhere

From~\eqref{desc_eq:1},~\eqref{desc_ineq:1},~\eqref{desc_ineq:2} and~\eqref{ineq:4}, we have the additive terms in~\eqref{desc_eq:1} (excluding $\gamma$) bounded by $o(\rho p_iq_i)$, hence we can write $g_i$ as $g_i = \rho p_iq_i(\lambda_i\cAi - \cAi^*) + o(\rho p_iq_i)$. Moreover, $\cAi$ is $2\delta$-close to
$\cAi^*$, then $\lambda_i = \inprod{\cAi}{\cAi^*} \geq 1-\delta
\approx 1$. Therefore, the update rule $g_i$ approximately aligns with the desired direction $\cAi
- \cAi^*$, which leads to the descent property argued in the next section.

\subsection{Descent property of $g_i^s$}

We now prove:
\begin{Lemma}
  \label{lm_correl_gi}
  The update $g_i^s$ is correlated with the desired direction $\cAi^s - \cAi^*$; that is, 
  $$\inprod{g_i^s}{\cAi^s - \cAi^*} \geq \rho p_iq_i(2 - \zeta^2) \norm{\cAi^s - \cAi^*}^2
+ \frac{1}{8\rho p_iq_i}\norm{g_i}^2 - \frac{\epsilon^2}{4\rho p_iq_i},$$
for $\zeta = 1 + 2\frac{1 - \rho}{\rho}\norm{A^*}_{\max} = 1 + o(1)$ and $\epsilon =
O(k^2/n^2)$. 

\end{Lemma}

\proof We prove this lemma by mainly using the results in the above
section. We first rewrite $g_i$ in
Equation~\eqref{desc_eq:1} in terms of the desired update direction
$\cAi^s - \cAi^*$ and everything else. For simplicity,  we omit the
superscript $s$ and $2\alpha = \rho p_iq_i$ throughout the proof. We have: 
\begin{align}
  \label{desc_eq:2}
  g_i &= \rho p_iq_i(\lambda_i\cAi \begin{aligned}[t] &- \cAi^*) + \rho p_i\cA{-i}\diag(q_{ij})\cA{-i}^T\cAi^* + (1 - \rho)p_iq_i\diag(\cAi^*\circ \cAi)\cAi \nonumber \\
        &+ (1-\rho)\sum_{j\neq i}p_iq_{ij}\diag(\cAi^*\circ \cAj)\cAj ~\pm \gamma
        \end{aligned} \\
      &= 2\alpha(A_i-A_i^*) + v, 
\end{align}
in which $v$ has the form:
\begin{align*}
v &= 2\alpha(\lambda_i - 1)\cAi + 2\alpha\frac{1 - \rho}{\rho} \diag(\cAi^*\circ \cAi)\cAi \\
&~~~+ 2\rho p_i\cA{-i}\diag(q_{ij})\cA{-i}^T\cAi^* + 2(1 - \rho)p_i\sum_{j\neq i}q_{ij}\diag(\cAi^*\circ \cAj)\cAj  ~\pm \gamma.
\end{align*}

First, we bound $\norm{v}$ in terms of $\norm{\cAi - \cAi^*}$. Since $\cAi$ is $\delta$-close to $\cAi^*$ and both
have unit norm, then $\norm{2\alpha(\lambda_i - 1)\cAi}  = \alpha
\norm{ \cAi - \cAi^*}^2 \leq \alpha \norm{\cAi - \cAi^*}$. 
Along with the bound of the second term obtained in~\eqref{desc_ineq:2}, we have 
\begin{equation}
  \label{desc_ineq:4}
  \norm{v}  \leq \alpha \left(1 + 2\frac{1 - \rho}{\rho}O(1/\sqrt{n}) 
  \right)\norm{\cAi - \cAi^*} + \epsilon = \alpha \zeta \norm{\cAi - \cAi^*} + \epsilon,
\end{equation}
where $\epsilon = \norm{2\rho p_i\cA{-i}\diag(q_{ij})\cA{-i}^T\cAi^* + 2(1 - \rho)p_i\sum_{j\neq i}q_{ij}\diag(\cAi^*\circ \cAj)\cAj \pm \gamma}= O(\rho k^2/m^2) + O((1-\rho) k^2/m^2) = O(k^2/m^2)$ due to~\eqref{desc_ineq:1} and~\eqref{ineq:4}. Here, $\zeta$ denotes the factor inside the parentheses. 

Now, we look at the correlation of $g_i$ and $\cAi - \cAi^*$ from~\eqref{desc_eq:2}:
\begin{align}
  \inprod{2g_i}{\cAi - \cAi^*} = 4\alpha \norm{\cAi - \cAi^*}^2 + \inprod{2v}{\cAi - \cAi^*}.
  \label{desc_eq:3}
\end{align}
Moreover, squaring both sides of~\eqref{desc_eq:2} and re-arranging
leads to
\begin{align}
  2\inprod{v}{ \cAi - \cAi^*} &= \frac{1}{2\alpha}\norm{g_i}^2 - 2\alpha \norm{\cAi - \cAi^*}^2 - \frac{1}{2\alpha}\norm{v}^2 \nonumber \\
                                      &\geq \frac{1}{2\alpha}\norm{g_i}^2 - 2\alpha\norm{\cAi - \cAi^*}^2 - \alpha\zeta^2 \norm{\cAi - \cAi^*}^2 - \frac{\epsilon^2}{\alpha},
\label{desc_ineq:5}
\end{align}
where in the last step we have used the Cauchy-Schwarz inequality: 
$$\norm{v}^2 \leq 2(\alpha^2\zeta^2 \norm{\cAi - \cAi^*}^2 + \epsilon^2),$$ 
applied to the right hand side of~\eqref{desc_ineq:4}.

Expressions~\eqref{desc_eq:3} and~\eqref{desc_ineq:5} imply that
$$ \inprod{2g_i}{\cAi - \cAi^*} \geq \alpha(2 - \zeta^2) \norm{\cAi - \cAi^*}^2
+ \frac{1}{2\alpha}\norm{g_i}^2 - \frac{\epsilon^2}{\alpha}.$$
Since $(1 -\rho)/\rho \leq k \leq O(\rho\sqrt{n}/\log n)$ and $m =
O(n)$, then $1 < \zeta^2 < 2$. Besides, we have $p_i = \Theta(k/m)$ and
$q_i = \Theta(1)$, then $\alpha = (1/2)\rho p_iq_i = \Theta(\rho
k/m)$, and $\epsilon^2/\alpha = O(k^3/\rho m^3)$ we have lower bound on the gradient. This is equivalent to saying that $g_i^s$
is $(\Omega(k/m), \Omega(m/k), O(k^3/\rho m^3))$-correlated with the true
solution $\cAi^*$ (see~\citep{arora-colt15}.) \qed

\proof[Proof of Theorem \ref{thm_desc_inf_sample}] Having argued the correlation of $g_i^s$ and $\cAi - \cAi^*$, we apply Theorem 6 in~\citep{arora-colt15} to obtain the descent stated in Theorem \ref{thm_desc_inf_sample}. Next, we will establish the nearness for the update at step $s$. \qedhere

\subsection{Nearness}
The final step in analyzing Algorithm~\ref{alg_descent} is to show that
the nearness of $A^{s+1}$ to the ground truth $A^*$ is maintained
after each update. Clearly, $A^{s+1}$ is columnwise close to $A^*$,
which follows from Theorem~\ref{thm_desc_inf_sample}. The final step
is to make sure that $\norm{A^{s+1} - A^*} \leq 2\norm{A^*}$ holds
true.

\begin{Lemma}
  \label{lm_nearness}
  Provided that $A^s$ is $(\delta, 2)$-near to $A^*$ and that the probability $\rho$ is a constant of $n$, then
  $\norm{A^{s+1} - A^*} \leq 2\norm{A^*}$.
\end{Lemma}
\proof Notice from the update that $A^{s+1} - A^* = A^s -
  A^* - \eta g^s$. Using the column-wise $g_i^s$ in~\eqref{desc_eq:1}, we have the matrix form for $g^s$ as
\begin{equation}
  \label{desc_eq:4}
  -\eta g^s = -\eta g^s_{\big \rvert_{\rho = 1}} - \eta(1-\rho)
  (A^{*}\circ A^s \circ A^s)\diag(p_iq_i) - \eta(1-\rho)Q \pm \eta\gamma,
\end{equation}
where $Q  \in \R^{n \times m}$ whose column $Q_i$ equals to $\sum_{j\neq i}p_iq_{ij}\diag(\cAj^*\circ \cAj)\cAj$. Since $\norm{A^s - A^*} \leq 2\norm{A^*}$, then to prove the lemma we need $\norm{\eta g^s} \leq o(\norm{A^*})$.~\citet{arora-colt15} have shown the same nearness
property for $\rho = 1$, \ie\  $\norm{\eta g^s_{\rvert_{\rho = 1}}} \leq o(\norm{A^*})$. We will show that the last two terms involving $1-\rho$ are negligible of $\norm{A^*}$.
From~\eqref{desc_ineq:3}, we have bound on each column $Q_i$ such that $\norm{Q_i} \leq O(\max_{j \neq i}q_{ij})$. Then,
\begin{equation*}
  \norm{Q} \leq \norm{Q}_F \leq \sqrt{m}\max_i\norm{Q_i} \leq O(\max_{j \neq i}q_{ij}\sqrt{m}) = O( k^2/m\sqrt{m}).
\end{equation*}
Moreover, $\eta =\Theta(m/\rho k)$ and $k \leq O^*(\rho\sqrt{n}/\log n)$
, therefore
\begin{equation*}
  \eta(1-\rho)\norm{Q} \leq O\Bigl(\frac{(1-\rho)k}{\rho \sqrt{m}}\Bigr) = o(1) 
\end{equation*}
We now bound the term $\eta(1-\rho)
(A^{*}\circ A^s \circ A^s)\diag(p_iq_i)$ using the column-wise upper bound in~\eqref{desc_ineq:2}. More specifically,
\begin{equation*}
  \norm{\eta(1-\rho) (A^{*}\circ A^s \circ A^s)\diag(p_iq_i)} \leq \sqrt{m}\norm{\eta(1-\rho)p_iq_i\diag(\cAi^*\circ \cAi)\cAi} \leq
O(\frac{m}{\rho k}(1-\rho)p_iq_i\delta\sqrt{m/n}) \leq o(1)
\end{equation*}
for a constant $\rho$ independent of $n$, $p_iq_i = \Theta(k/m)$ and $m = O(n)$. Put together, we complete the proof of Lemma~\ref{lm_nearness}. \qedhere

\section{Analysis of Algorithm~\ref{alg_init}}
\label{app:init}

\proof[Proof of Lemma~\ref{lm_Muv}] Recall the distributional properties of $x^*$ that
$x^*_i$'s are conditionally independent given $S = \supp(x^*)$ and the
summary statistics are $\E[x_i^{*4}|i\in S] = c_i \in (0, 1)$, $\E[x_i^{*2}|i\in S] =
1$, $q_i = \Prob[i \in S]$ and $q_{ij} = \Prob[i, j \in S]$.
\begin{align*}
  M_{u, v} &= \frac{1}{\rho^4}\E[\inprod{y}{u}\inprod{y}{v}yy^T] =
    \frac{1}{\rho^2}\E\bigl[\inprod{x^*}{\beta}\inprod{x^*}{\beta'} \rAG^*x^*x^{*T} \rAG^{*T} \bigr] \\
  &= \frac{1}{\rho^2}\E_{\Gamma}\E_{x^*}\Bigl[\sum_{i \in S}\beta_ix_i^*\sum_{i
    \in S}\beta'_ix_i^*\sum_{i, j \in S}x_i^*x_j^*\AG{i}^*\AG{i}^{*T}\Bigr] \\
  &= \frac{1}{\rho^2}\sum_{i\in
    [m]}q_ic_i\beta_i\beta'_i\E_{\Gamma}[\AG{i}^*\AG{i}^{*T}] + \frac{1}{\rho^2}\sum_{i,j \in [m], j \neq i}
    q_{ij}\beta_i\beta'_i\E_{\Gamma}[\AG{j}^*\AG{j}^{*T}] +
    2q_{ij}\beta_i\beta'_j\E_{\Gamma}[\AG{i}^*\AG{j}^{*T}],
\end{align*}
We continue calculating the expectations over $\Gamma$. All of those
terms are of the same form:
\begin{align*}
  \E_{\Gamma}[\AG{i}^*\AG{j}^{*T}] = \rho(1-\rho) \diag(\cAi^*\circ \cAj^*) + \rho^2 \cAi^*\cAj^{*T}.
\end{align*}
Plug in this expression into $M_{u, v}$ to have, 
\begin{align*}
  M_{u, v} &= \sum_{i \in U \cap V} q_ic_i\beta_i\beta'_i\cAi^*\cAi^{*T} + \sum_{i \notin U \cap V} q_ic_i\beta_i\beta'_i\cAi^*\cAi^{*T} +
             \sum_{j \neq i}q_{ij}\beta_i\beta'_i\cAj^*\cAj^{*T} +  2 \sum_{j \neq i}q_{ij}\beta_i\beta'_j\cAi^*\cAj^{*T} \\
           &+ \frac{1-\rho}{\rho}\sum_{i\in [m]}
    q_i\beta_i\beta'_i\diag(\cAi^*\circ \cAi^*) + \frac{1-\rho}{\rho} \sum_{j \neq i}
    q_{ij}\beta_i\beta'_i\diag(\cAj^*\circ \cAj^*) +
    2q_{ij}\beta_i\beta'_j\diag(\cAi^*\circ \cAj^*) \\
           &= \sum_{i \in U \cap V} q_ic_i\beta_i\beta'_i\cAi^*\cAi^{*T} 
    +~\text{perturbation terms},
\end{align*}
where all the terms except $\sum_{i \in U \cap V} q_ic_i\beta_i\beta'_i\cAi^*\cAi^{*T}$ are expected to be small enough. When $\rho = 1$, then $M_{u, v}$ simply includes the first four terms, which is exactly the weighted matrix studied in~\citep{arora-colt15} for regular sparse coding. We will adapt bounds for these terms that now depend on $\rho$. First of all, for $i
\notin U \cap V$ assume $\alpha_i = 0$, using Claim~\ref{cl_beta_estimate} and $\abs{\alpha_i'} \leq O(\log n)$ we have $\abs{\beta_i\beta_i'} \leq \abs{(\beta_i - \alpha_i)(\beta_i' - \alpha_i')} + \abs{\beta_i\alpha_i'} \leq O^*(1/\log n)$, then 
\begin{equation}
  \label{init_ineq:1}
  \norm[\Big]{\sum_{i \notin U \cap V} q_ic_i\beta_i\beta'_i\cAi^*\cAi^{*T}} \leq O^*(k/m\log n),
\end{equation}
for $q_i = \Theta(k/m)$. For next two perturbation terms, recall from Claim~\ref{cl_beta_estimate} $\beta$ and $\beta'$ has norms bounded by $O(\sqrt{k}\log n/\rho)$ and $q_{ij} = \Theta(k^2/m^2)$. We again use the results from~\citep{arora-colt15} to get
\begin{equation}
  \label{init_ineq:2}
  \norm[\Big]{\sum_{j \neq i}
     q_{ij}(\beta_i\beta'_i\cAj^*\cAj^{*T} +
    2\beta_i\beta'_j\cAi^*\cAj^{*T})} \leq O\Bigl(\frac{k^3\log^2n}{\rho^2m^2}\Bigr).
\end{equation}
%
Now, we will handle the terms involving the diagonal matrices as follows,
\begin{align}
  \label{init_ineq:3}
  \norm[\Big]{\sum_{i\in [m]} q_i\beta_i\beta'_i\diag(\cAi^*\circ \cAi^*)} &= \max_{j\in [n]}\abs[\big]{\sum_{i\in [m]}q_i\beta_i\beta'_iA_{ji}^{*2}} \leq \max_{i, j}(q_iA_{ji}^{*2})\abs[\big]{\sum_{i\in m}\beta_i\beta'_i} \nonumber \\
                                                                     &\leq
           \max_{i}q_i\norm{A^*}_{\max}^2\norm{\beta}\norm{\beta'} = O\Bigl(\frac{k^2\log^2n}{\rho^2mn}\Bigr)
\end{align}
because of the fact that $\norm{A^*}_{\max} \leq O(1/\sqrt{n})$. Similarly, we also have the same bound for the below term
\begin{align}
  \label{init_ineq:4}
  \norm[\Big]{\sum_{j \neq i} q_{ij}\beta_i\beta'_i\diag(\cAj^*\circ
  \cAj^*)} &= \max_{l\in [n]}\abs[\big]{\sum_{j \neq
             i}q_{ij}\beta_i\beta'_iA_{lj}^{*2}} = \max_{l\in [n]}\abs[\big]{\sum_i\beta_i\beta'_i\sum_{j \neq i}q_{ij}A_{lj}^{*2}} \nonumber \\
           &\leq \max_{i, l}(\sum_{j \neq i}q_{ij}A_{lj}^{*2})\abs[\big]{\sum_{i\in m}\beta_i\beta'_i} \leq \max_{i, l}(\sum_{j \neq
             i}q_{ij}A_{lj}^{*2})\norm{\beta}\norm{\beta'} \nonumber\\
           &= O\Bigl(\frac{k^2\log^2n}{\rho^2mn}\Bigr),
\end{align}
where we used $\sum_{j \neq i}q_{ij}A_{lj}^{*2} \leq \max_{i \neq j}q_{ij}\norm{\rAl^*}^2 \leq O(k^2/mn)$ since $\norm{\rAl^*} \leq \norm{A^*} \leq O(\sqrt{m/n})$.

We bound the last term using a result from~\citep{nguyen-aaai18} (proof of Claim 4) that $\sum_{j \neq i} q_{ij} \beta_i\beta'_jA^*_{li}A^*_{lj} = \rAl^{*T}Q_\beta\rAl^*$ where $(Q_\beta)_{ij} = q_{ij} \beta_i\beta'_j$ for $i \neq j$ and $(Q_\beta)_{ij} = 0$ for $i = j$, so
\begin{align*}
   \abs{\rAl^{*T}Q_\beta\rAl^*} \leq \norm{Q_\beta}\norm{\rAl^*}^2 \leq  \norm{Q_\beta}_F \norm{A^*}^2_{1,2},
\end{align*}
Moreover, $ \norm{Q_\beta}^2_F = \sum_{i\neq j}  q_{ij}^2 \beta_i^2(\beta'_j)^2 \leq (\max_{i\neq j}q_{ij}^2)\sum_i \beta_i^2\sum_j(\beta'_j)^2 \leq (\max_{i\neq j}q_{ij}^2) \norm{\beta}^2 \norm{\beta'}^2$, then
\begin{align}
  \label{init_ineq:5}
  \norm[\Big]{\sum_{j \neq i}q_{ij}\beta_i\beta'_j\diag(\cAi^*\circ \cAj^*} &= \max_{l\in [n]}\abs[\big]{\sum_{j \neq
  i}q_{ij}\beta_i\beta'_jA_{li}^*A_{lj}^*} = \max_{l\in [n]}\abs{\rAl^{*T}Q_{\beta}\rAl^*} \nonumber \\
                                                                            &\leq (\max_{i\neq j}q_{ij}^2) \norm{\beta}^2 \norm{\beta'}^2 \leq O\Bigl(\frac{k^2\log^2n}{\rho^2m^2}\Bigr).
\end{align}

Since $(1-\rho)/\rho \leq k$ and $m = O(n)$, then~\eqref{init_ineq:2},
~\eqref{init_ineq:3}, ~\eqref{init_ineq:4} and~\eqref{init_ineq:5} are
all bounded by $O\Bigl(\frac{k^3\log^2n}{\rho^2mn}\Bigr)$. Besides, we
know that $ k \leq O^*(\frac{\rho\sqrt{n}}{\log n})$, then all the
perturbation terms are bounded by $O^*(k/m\log n)$. We have finished
the proof of Lemma~\ref{lm_Muv}. \qedhere


\section{Sample Complexity}
\label{app:concen}


In this section, we give concentration bounds for the finite-sample
estimates $\ghat^s$ and $\Mhuv$ and prove
Theorem~\ref{main_thm_desc} and Theorem~\ref{main_thm_init}
. We employ the same technique used in~\citep{arora-colt15}, which
basically apply Bernstein inequalities for proper vector and matrix random
variables. The inequality is generally stated in the following lemma.

\begin{Lemma}
    \label{lm_bernstein_ineq}
 Suppose that $Z^{(1)}, Z^{(2)}, \dots, Z^{(p)}$ are $p$ \iid\ samples
 drawn from some distribution $\mathcal{D}$ such that $\E[Z^{(j)}] = 0$, $\norm{Z^{(j)}} \leq \Rad$ almost surely and $\norm{\E[Z^{(j)}(Z^{(j)})^T} \leq \sigma^2$ for each $j$, then 
 \begin{equation}
   \frac{1}{p}\norm[\Big]{\sum_{j=1}^p Z^{(j)}} \leq \Otilde\biggl(\frac{\Rad}{p} + \sqrt{\frac{\sigma^2}{p}}\biggr)
 \end{equation}
holds with probability $1-n^{-\omega(1)}$. 
\end{Lemma}

In order to apply the above inequality, we need bounds on the random
variable $Z$ and its covariance. However, these
quantities are not bounded almost surely, and hence we use the common trick of analyzing a truncated
version of $Z$ to
overcome this issue. Lemma~\ref{lm_trunc_trick} provides sufficient
conditions for the truncation trick to work

\begin{Lemma} [\citet{arora-colt15}]
  \label{lm_trunc_trick}
  Suppose a random variable $Z$ satisfies $\Prob[\norm{Z} \geq \Rad(\log(1/\rho))^C] \leq \rho$ for some constant $C > 0$, then

\begin{enumerate}
\item If $p = n^{O(1)}$, it holds that $\norm{Z^{(j)}} \leq \Otilde(\Rad)$ for each $j$ with probability $1 - n^{-\omega(1)}$.
\item $\norm{\E[Z\1_{\norm{Z} \geq \Omgtilde(\Rad)}]} = n^{-\omega(1)}$.
\end{enumerate}

\end{Lemma}

Note that there is a slight abuse of notation here: the constant $C$ and $\rho$ are only used in the context of
the above lemma and are not related to those
used in our generative model. Since the random components in $\ghat$
and $\Mhuv$ are products of sub-Gaussian random variables, we can apply
Lemma~\ref{lm_bernstein_ineq} and Lemma~\ref{lm_trunc_trick} to show the concentration of $\frac{1}{p}
\sum_{i=1}^p Z^{(j)}(1 - \1_{\norm{Z^{(j)}} \geq \Omgtilde(\Rad)})$,
then conclude about the concentration of $\frac{1}{p} \sum_{i=1}^p
Z^{(j)}$ likewise.

In bounding $\norm{\E[{Z}{Z}^T(1 - \1_{\norm{{Z}} \geq
    \Omgtilde(\Rad))}]}$, we sometimes 
need to take bounds of some random terms out of the expectation. In such
case, the following lemma is often useful.

\begin{Lemma}[\citet{nguyen-aaai18}]
  \label{aux_lm_pull_out_prob_bound}
  Suppose a random variable $\tilde{Z}\tilde{Z}^T = aT$ where $a \geq 0$ and $T$ is positive semi-definite.
 Suppose
 $\Prob[a \geq \Aupb]=n^{-\omega(1)}$
 and $\Bupb>0$ is a constant.
 Then,
$$\norm{\E[\tilde{Z}\tilde{Z}^T(1 - \1_{\norm{\tilde{Z}} \geq \Bupb})]} \leq \Aupb\norm{\E[T]} + O( n^{-\omega(1)})$$
\end{Lemma}

Other details of these auxiliary lemmas can be found
in~\citep{arora-colt15, nguyen-aaai18}.

\subsection{Sample Complexity of Algorithm \ref{alg_descent}}
\subsubsection{Proof of Theorem~\ref{main_thm_desc}}


We start by using two key auxiliary lemmas for the concentration of $\ghat$, both column-wise as well as for the whole matrix.

\begin{Lemma}
At iteration $s$ of Algorithm~\ref{alg_descent}, suppose
that $A^s$ is $(\delta_s,
2)$-near to $A^*$. Then $\norm{\ghat^s_i - g_i^s} \leq O(k/m)\cdot( o(\delta_s) +
O(\epsilon_s))$ with high probability for $\delta_s = O^*(1/\log n)$ and $\epsilon_s = O(\sqrt{k/n})$ when $p = \Omgtilde(m)$.
\label{lm_concentration_gi}
\end{Lemma}

\begin{Lemma}
  \label{lm_nearness_finite_sample}
  If $A^s$ is $(\delta_s, 2)$-near to $A^*$ and number of samples used in step $s$ is $p=\Omgtilde(mk)$, then with high probability $\norm{A^{s+1} - A^*} \leq 2\norm{A^*}$.
\end{Lemma}
While the proof of Lemma~\ref{lm_concentration_gi} is provided below, Lemma~\ref{lm_nearness_finite_sample} directly follows from Lemma 42 in~\citet{arora-colt15} and the number of samples being $\Omgtilde(mk)$.
\proof[Proof of Theorem~\ref{main_thm_desc}] We can write $\ghat_i^s$
as
\begin{equation*}
  \ghat_i^s = g_i^s + ( \ghat_i^s - g_i^s) = g_i^s + O(k/m)\cdot( o(\delta_s) +
O(\epsilon_s))
\end{equation*}
with high probability;  then argue that $\ghat_i^s$ is correlated with $\cAi - \cAi^*$ with
high probability from Lemma~\ref{lm_correl_gi}. The descent property follows
directly as Theorem~\ref{thm_desc_inf_sample} except that we have the expected
$\inprod{\ghat_i^s}{\cAi - \cAi^*}$ on the right hand side. The overall sample complexity is $\Otilde(mk)$, which combines the complexities of having descent and maintaining nearness. \qedhere

\subsubsection{Proof of Lemma~\ref{lm_concentration_gi}}

Notice that $\ghat^s_i$ is a sum of $p$ random vectors of the form
$(\mathcal{P}_{\Gamma}(Ax) - y)\sgn(x_i)$. We will show the
concentration of $\ghat^s_i$ by applying the Bernstein
inequality on $Z \triangleq (\mathcal{P}_{\Gamma}(Ax) - y)\sgn(x_i)$. Nevertheless, the inequality does not give a sharp
bound for such sparse $Z$, so we instead consider $Z \triangleq (\mathcal{P}_{\Gamma}(Ax) - y)\sgn(x_i) | i \in S$,
with $S = \supp(x^*)$ and $x = \thres_{C/2}(A^Ty)$. 

\begin{Claim}
  \label{cl_concentration_sparse_Zr}
  Suppose that $Z^{(1)}, Z^{(2)}, \dots, Z^{(N)}$ are \iid\ samples of the random variable $Z = \mathcal{P}_{\Gamma}(y - Ax)\sgn(x_i) | i \in S$. Then,
 \begin{equation}
   \norm[\Big]{\frac{1}{N}\sum_{j=1}^N Z^{(j)} - \E[Z]} \leq o(\delta_s) + O(\epsilon_s)
 \end{equation}
holds with probability when $N = \Omgtilde(k)$,
$\delta_s = O^*(1/\log n)$ and $\epsilon_s = O(\sqrt{k/n})$.
\end{Claim}

\proof[Proof of Lemma~\ref{lm_concentration_gi}] The lemma is easily
proved by applying Claim~\ref{cl_concentration_sparse_Zr}. For the
reader, we recycle the proof of Lemma 43 in \cite{arora-colt15}.

Write $W = \{j: i \in \supp(x^{*(j)})\}$ and $N = |W|$, then express $\ghat_i$ as 
$$\ghat_i = \frac{N}{p}\frac{1}{N} \sum_{j}(\mathcal{P}_{\Gamma}(Ax^{(j)}) - y^{(j)})\sgn(x_i^{(j)}),$$
where $\frac{1}{|W|} \sum_{j}(\mathcal{P}_{\Gamma}(Ax^{(j)}) - y^{(j)})\sgn(x_i^{(j)})$ is distributed as $\frac{1}{N}\sum_{j=1}^N Z^{(j)}$ with $N = |W|$. Note that $\E[(\mathcal{P}_{\Gamma}(Ax) - y)\sgn(x_i)] = \E[(\mathcal{P}_{\Gamma}(Ax) - y)\sgn(x_i)\1_{i\in S}] = \E[Z]\Prob[i \in S] = q_i\E[Z]$ with $q_i = \Theta(k/m)$. Following Claim \ref{cl_concentration_sparse_Zr}, we have 

$$\norm{\ghat^s_i - g_i^s} \leq O(k/m)\norm[\Big]{\frac{1}{N}\sum_{j=1}^N Z^{(j)} - \E[Z]} \leq  O(k/m)\cdot( o(\delta_s) + O(\epsilon_s)),$$
holds with high probability as $p = \Omega(mN/k)$. Substituting $N$ in Claim \ref{cl_concentration_sparse_Zr}, we obtain the results in Lemma \ref{lm_concentration_gi}. \qedhere

\proof[Proof of Claim \ref{cl_concentration_sparse_Zr}] To prove it,
we need to bound $\norm{Z}$ and its variance (Lemma~\ref{cl_norm_bound_ghi} and Lemma~\ref{cl_bound_variance_ghi}), then we can apply the
Bernstein inequality in Lemma~\ref{lm_bernstein_ineq}.

\begin{Claim}
  \label{cl_norm_bound_ghi}
  $\norm{Z} \leq
  \Otilde(\delta_s\sqrt{k} + \mu k/\sqrt{n})$ holds with high
  probability over the randomness of $y$.
\end{Claim}

\proof From the generative model and the support consistency of the
encoding step, we have
$$y = \PGam(A^*x^*) = \AG{S}^*x^*_S ~\text{and } x_S = \cAS^Ty = \cAS^T\AG{S}^*x^*_S$$ 
and plug the following quantities into the 
\begin{align*}
   y - \PGam(Ax) &= \AG{S}^*x_S^* - \AG{S}\cAS^T\AG{S}^*x^*_S \\
   &= (\AG{S}^* - \AG{S})x^*_S + \AG{S}(I_n - \cAS^T\AG{S}^*)x^*_S.
\end{align*}

By the fact that $x^*_S$ is sub-Gaussian and $\norm{Mw} \leq
\Otilde(\sigma_w\norm{M}_F)$ holds with high probability for a fixed
$M$ and a sub-Gaussian $w$ of entrywise variance $\sigma_w^2$, we have
$$\norm{(\PGam(Ax) - y)\sgn(x_i) | i \in S} \leq \Otilde(\norm{\AG{S}^* - \AG{S}}_F + \norm{\AG{S}(I_k - \cAS^T\cA{S}^*)}_F).$$

Now, we need to bound those Frobenius norms. The first quantity is easily bounded as 
\begin{equation}
  \label{ineq_C2.1}
  \norm{\AG{S}^* - \AG{S}}^2_F = \sum_{i \in S} \norm{\AG{i} - \AG{i}^*}^2 \leq \delta_s^2k
\end{equation}
due to the $\delta$-closeness of $A$ and $A^*$. This leads to
$\norm{\AG{S}^* - \AG{S}}_F \leq \delta_s\sqrt{k}$ \whp\ To handle the other two, we use the fact that $\norm{UV}_F \leq \norm{U} \norm{V}_F$. For the second term, we have
$$\norm{\AG{S}(I_k - \cAS^T\cAS^*)}_F \leq \norm{\AG{S}}\norm{(I_k - \cAS^T\cAS^*)}_F, $$
where $\norm{\AG{S}} \leq \norm{\rAG} \leq O(1)$ due to the nearness.

The second part is rearranged to take advantage of the closeness and incoherence properties:
\begin{align*}
  \label{ineq_C2.2}
   \norm{I_k - \cAS^T\cAS^*}_F  &\leq \norm{I_k - \cAS^{*T}\cAS^* - (\cAS - \cAS^*)^T\cAS^*}_F \\
        &\leq \norm{I_k - \cAS^{*T}\cAS^*}_F + \norm{(\cAS - \cAS^*)^T\cAS^*}_F \\
  &\leq \norm{I_k - \cAS^{*T}\cAS^*}_F + \norm{\cAS^*}\norm{\cAS - \cAS^*}_F \\
        &\leq \mu k/\sqrt{n} + O(\delta_s\sqrt{k}),
\end{align*}
where we have used $\norm{I_k - \cAS^{*T}\cAS^*}_F \leq \mu k/\sqrt{n}$ because of the $\mu$-incoherence of $A^*$,  $\norm{\cAS - \cAS^*}_F \leq \delta_s\sqrt{k}$ in \eqref{ineq_C2.1} and $\norm{\cAS^*} \leq \norm{A^*} \leq O(1)$. Accordingly, the second Frobenius norm is bounded by 
\begin{equation}
  \label{ineq_C2.31}
  \norm{\AG{S}(I_k - \cAS^T\cAS^*)}_F \leq  O\bigl(\mu k/\sqrt{n} + \delta_s\sqrt{k}\bigr).
\end{equation}

\begin{Claim} 
  \label{cl_bound_variance_ghi}
$\E[\norm{Z}^2] \leq O(\delta_s^2k + k^2/n)$ holds with $\delta_s = O^*(1/\log n)$.
\end{Claim}

\proof
In the following proofs, we use $x_S^*$ to mean a vector of size $k$ obtained by selecting entries in $S$. Using the fact that $E[x^*_Sx^{*T}_S] = I_k$, we can expand the expectation $\E[\norm{Z}^2]$ as follows,
\begin{align*}
\E[\norm{\PGam(y-Ax)\sgn(x_i)}^2|i\in S] &= \E[\norm{(\AG{S}^* - \AG{S}\cAS^T\cAS^*)x^*_S}^2] \\
                                         &= \E[\norm{\AG{S}^* -
                                        \AG{S}\cAS^T\cAS^*}_F^2| i\in S] \\
                                         &\leq \E[\norm{(\AG{S}^* - \AG{S})}^2| i\in S] + \E[\norm{\AG{S}(I_k - \cAS^T\cAS^*)}^2| i\in S] \\
                                         &\leq \delta_s^2k + \E[\norm{\AG{S}(I_k - \cAS^T\cAS^*)}^2| i\in S].
\end{align*}
Here we have used the bound $\norm{(\AG{S}^* - \AG{S})}^2 \leq \delta_s^2k$ for the first term shown in the previous claim. For the second term, we notice that
\begin{align}
  \label{ineq_C2.5}
  \E[\norm{\AG{S}(I_k - \cAS^T\cAS^*)}_F^2|i \in S] &\leq \sup_S\norm{\AG{S}}^2\E[\norm{I_k - \cAS^T\cAS^*}_F^2|i \in S],
\end{align}
in which $\sup_S\norm{\AG{S}} \leq \norm{\rAG} \leq O(1)$. 
We will show that $\E[\norm{I_k - \cAS^T\cAS^*}_F^2|i \in S] \leq O(k\delta_s^2) + O(k^2/n)$ by recycling the proof from~\cite{arora-colt15}:
\begin{align*}
  &\E[\norm{I_k - \cAS^T\cAS^*}_F^2|i \in S] = \E[\sum_{j \in S}(1 - \cAj^T\cAj^*)^2 + \sum_{j \in S}\norm{\cAj^TA_{\bigcdot, -j}^*}^2|i \in S] \\
  &= \E[\sum_{j \in S}\frac{1}{4}\norm{\cAj - \cAj^*}^2] +  q_{ij}\sum_{j \neq i}\norm{\cAj^TA_{\bigcdot, -j}^*}^2 + q_i \norm{\cAi^TA_{\bigcdot, -i}^*}^2 + q_i \norm{A_{\bigcdot, -i}^T\cAi^*}^2,
\end{align*}
where $A_{\bigcdot, -i}$ is the matrix $A$ with the $i^{\textrm{th}}$ column removed, $q_{ij} \leq O(k^2/m^2)$ and $q_i \leq O(k/m)$. For any $j = 1, 2, \dots, m$,
\begin{align*}
  \norm{\cAj^TA_{\bigcdot, -j}^*}^2 &= \norm{\cAj^{*^T}A_{\bigcdot, -j}^* + (\cAj - \cAj^*)^TA_{\bigcdot, -j}^*}^2 \\
&\leq \sum_{l \neq j}\inprod{\cAj^*}{\cAl^*}^2 + \norm{(\cAj - \cAj^*)^TA_{\bigcdot, -j}^*}^2 \\
  &\leq \sum_{l \neq j}\inprod{\cAj^*}{\cAl^*}^2 + \norm{\cAj - \cAj^*}^2\norm{A_{\bigcdot, -j}^*}^2 \leq \mu^2 + \delta_s^2.
\end{align*}
The $\mu$-incoherence, $\delta$-closeness and the spectral norm of $A^*$ have been used in the last step. Similarly, we can bound $\norm{\cAi^TA_{\bigcdot, -i}^*}^2$ and $\norm{A_{\bigcdot, -i}^T\cAi^*}^2$. As a result,
\begin{align}
  \label{ineq_C2.6}
  \E[\norm{I_k - \cAS^T\cAS^*}_F^2|i \in S] \leq O(k\delta_s^2) + O(k^2/n).
\end{align}

Combining~\eqref{ineq_C2.5} and~\eqref{ineq_C2.6}, we have shown that the covariance is bounded by: $\sigma^2 = O(\delta_s^2k + k^2/n)$. \qedhere

Having had $R = \Otilde(\delta_s\sqrt{k} + \mu k/\sqrt{n})$ and  $\sigma^2 = O(\delta_s^2k + k^2/n)$ in Claims \ref{cl_norm_bound_ghi} and \ref{cl_bound_variance_ghi}, we are now ready to apply truncated Bernstein inequality to the random variable $Z^{(j)}(1-1_{\norm{Z^{(j)}} \geq \Omega(R)})$, leading to the concentration of $\frac{1}{N}\sum_{j=1}^N Z^{(j)}$. More precisely,

\begin{equation*}
  \norm[\Big]{\frac{1}{N} \sum_{i=1}^N Z^{(j)} - E[Z]} \leq
  \Otilde\Bigl(\frac{\Rad}{N}\Bigr) +
  \Otilde\biggl(\sqrt{\frac{\sigma^2}{N}}\biggr) = o(\delta_s) +
  O(\sqrt{k/n})
\end{equation*}
holds with high probability when $N = \Omgtilde(k)$. As such, we finished the proof of Claim~\ref{cl_concentration_sparse_Zr}. \qedhere


\subsection{Sample Complexity of Algorithm \ref{alg_init}}

In the next proofs, we argue the concentration inequality for $\Mhuv$ computed in Algorithm \ref{alg_init}, which is the empirical average over \iid\ samples of $y$, then prove Theorem~\ref{main_thm_init}. We note that while $u$ and $v$ are fixed for one iteration, they are random. The (conditional) expectations contain randomness from $u$ and $v$, hence in some high probability statement, we refer it to the randomness of $u, v$.

\begin{Lemma}
  \label{lm_concentration_ehat_Mhat}
  Consider Algorithm \ref{alg_init} in which $p$ is
  the given number of incomplete samples. For any pair of full samples $u$ and $v$,  with
  high probability $\norm{\Mhuv - \Muv} \leq
  O^*(k/m\log n)$ when $p = \Omgtilde(mk/\rho^4)$.
\end{Lemma}

\subsubsection{Proof of Lemma \ref{lm_concentration_ehat_Mhat}} Consider a random matrix variable $Z \triangleq \inprod{y}{u}\inprod{y}{v} yy^T$. We have $\Mhuv = \frac{1}{p}\sum_{i=1}^pZ^{(i)}/\rho^4$. To give a tail bound for $\norm{\Mhuv - \Muv}$, all we need is derive are an upper norm bound
$\Rad$ of the matrix random variable $Z$ and its variance, then apply Bernstein inequality. These following claims provide bounds for $\norm{Z}$ and $\norm{\E[ZZ^T]}$.

\begin{Claim}
  \label{cl_bound_y}
  $\norm{y} \leq \Otilde(\sqrt{k})$ and $\abs{\inprod{y}{u}} \leq \Otilde(\sqrt{k})$ hold with high probability (over random samples $u$ and $v$.)
\end{Claim}
\proof Under the generative model where $S = \supp(x^*)$, we have
$$\norm{y} = \norm{\AG{S}^*x_S^*} \leq \norm{\AG{S}^*x_S^*} \leq \norm{\AG{S}^*}\norm{x_S^*}. $$
From Claim \ref{cl_beta_estimate}, $\norm{x_S^*} \leq \Otilde(\sqrt{k})$ \whp\ In addition, $\norm{\AG{S}^*}  \leq \norm{A^*} \leq O(1)$. Therefore, $\norm{y} \leq \Otilde(\sqrt{k})$ \whp, which is the first part of the proof. To bound the second term, we write it as
$$\abs{\inprod{y}{u}} = \abs{\inprod{\AG{S}^*x_S^*}{u}} \leq \abs{\inprod{x_S^*}{\AG{S}^{*T}u}}.$$
Even though $u$ is fully observed sample, we can prove similarly that $\norm{u} \leq \Otilde(\sqrt{k})$ \whp\, which results in $\norm{\cAS^{*T}u} \leq \norm{\cAS^{*T}}\norm{u} \leq \Otilde(\sqrt{k})$ with high probability. Consequently,
$\abs{\inprod{y}{u}} \leq \Otilde(\sqrt{k})$ \whp,
and we finish the proof of the claim. \qed

\begin{Claim}
  \label{cl_bound_Mh_A41}
$\norm{Z} \leq \Otilde(k^2)$ and $\norm{\E[ZZ^T]} \leq \Otilde(\rho^4k^3/m)$ hold with high probability.
\end{Claim}
\proof  First, it is obvious that $$\norm{Z} \leq \abs{\inprod{y}{u} \inprod{y}{v}} \norm{y}^2,$$
in which $\abs{\inprod{y}{u} \inprod{y}{v}} \leq \Otilde(k)$ and
$\norm{y}^2 \leq \Otilde(k)$ \whp\ (according to Claim \ref{cl_bound_y}). Clearly, $\norm{Z} \leq \Otilde(k^2)$ \whp\

For the second part, we use the auxiliary lemma~\ref{aux_lm_pull_out_prob_bound} to take out the bound of $\norm{Z}$. Specifically, we have just shown that $\norm{Z} \leq \Otilde(k^2)$ and $\inprod{y}{v}^2\norm{y}^2 \leq \Otilde(k^2)$, applying Lemma \ref{aux_lm_pull_out_prob_bound}:
\begin{equation*}
  \norm{\E[ZZ^T(1 - \1_{\norm{Z} \geq \Omgtilde(k^2)})]} \leq \Otilde(k^2)\norm{\E[\inprod{y}{u}^2 yy^T]} +
\Otilde(k^2)O(n^{-\omega(1)}) \leq \Otilde(k^2)\norm{\rho^4M_{u, u}}, 
\end{equation*}
where $M_{u,u}$ is the expected weighted covariance matrix defined in
Lemma~\ref{lm_Muv} for $u$ and $v = u$. From Lemma~\ref{lm_Muv} we have
$$M_{u,u} = \sum_{i}q_ic_i\beta_i^2\cAi^*\cAi^{*T} + ~\text{perturbation terms},$$
and the perturbation terms are all bounded by $O^*(k/m\log n)$ whereas
\begin{equation*}
  \norm{\sum_{i}q_ic_i\beta_i^2\cAi^*\cAi^{*T}} = \norm{A^*\diag(q_ic_i\beta_i)A^{*T}}\leq (\max_{i}q_ic_i\beta_i^2)\norm{A^*}^2 \leq \Otilde(k/\rho m)\norm{A^*}^2 \leq \Otilde(k/m)
\end{equation*}
\whp\ since $\abs{\beta_i} \leq \log n$ \whp\ Finally, the variance bound is $\Otilde(\rho^4k^3/m)$ \whp\ \qed

Then, applying Bernstein inequality in Lemma \ref{lm_bernstein_ineq} to the truncated version of $Z$ with $\Rad = \Otilde(k^2)$ and variance $\sigma^2 = \Otilde(\rho^4k^3/m)$ and obtain the concentration for the full $Z$ to get 
$$\norm{\Mhuv - \Muv} \leq \frac{\Otilde(k^2)}{\rho^4p} + \frac{1}{\rho^4}\sqrt{\frac{\Otilde(\rho^4k^3/m)}{p}} \leq O^*(k/m\log n)$$
\whp\ when the number of samples is $p = \Omgtilde(mk/\rho^4)$. We finish the proof of Lemma \ref{lm_concentration_ehat_Mhat}. \qedhere

\subsubsection{Proof of Theorem~\ref{main_thm_init}}
We can write the empirical estimate $\Mhuv$ in term of its expectation $\Muv$ as 
$$\Mhuv = q_ic_i\beta_i\beta'_i\cAi^*\cAi^{*T} + ~\text{perturbation terms} + (\Mhuv - \Muv),$$
and the new term $\Mhuv - \Muv$ can be considered an additional perturbation with the same magnitude $O^*(k/m\log n)$ in spectral norm. As a consequence, as $u$ and $v$ share a unique element in their code supports, the top singular vectors of $\Mhuv$ is $O^*(1/\log n)$ -close to $\cAi^*$ with high probability using $p = \Otilde(mk/\rho^4)$ partial samples.

Each vector added to the list $L$ in Algorithm~\ref{alg_init}
is close to one of the dictionary, then it must be the case that $A^0$ is $\delta$-close
to $A^*$. In addition, the nearness of $A^0$ to $A^*$ is guaranteed via
an appropriate projection onto the convex set $\mathcal{B} = \{A | A
~\text{close to } A^0 ~\text{and}~ \norm{A} \leq
2\norm{A^*} \}$.

Finally, using the result in~\citep{arora-colt15}, the number of full samples in $\mathcal{P}_1$ is $\Otilde(m)$ such that we can draw $u, v$ share uniquely and estimate all the $m$ dictionary atoms. Overall, the sample complexities of Algorithm~\ref{alg_init} are $\Otilde(m)$ full samples and $p = \Otilde(mk/\rho^4)$ partial samples. We finish the proof of Theorem \ref{main_thm_init}. \qedhere



\end{document}